\documentclass[final,3p,times]{elsarticle}
\usepackage{graphicx}
\usepackage{amssymb}
\usepackage{amsmath}

\usepackage{amsfonts}
\usepackage{amsthm}
\newtheorem*{remark}{Remark}
\usepackage{mathtools}

\usepackage{amsbsy}
\usepackage{physics} 

\usepackage{url}
\usepackage{bm}
\usepackage{algorithm}
\usepackage{algpseudocode}
\usepackage{lineno}
\usepackage{booktabs}
\usepackage{adjustbox}
\usepackage{xcolor}
\usepackage{subcaption}
\usepackage{lipsum}

\newcommand{\trp}{^{\mathsf{T}}}

\begin{document}

\begin{frontmatter}


\title{Gaussian process learning with flow map refinement for parameter estimation in dynamical systems}



\author[ustsaddress]{Yue Hao}
\ead{y.hao@usts.edu.cn}

\author[xjtluaddress]{Dongwei.Ye\corref{mycorrespondingauthor}}
\cortext[mycorrespondingauthor]{Corresponding authors}
\ead{Dongwei.Ye@xjtlu.edu.cn}

% \ead{c.brune@utwente.nl}

% \author[uvaaddress]{Mengwu Guo}
% \ead{m.guo@utwente.nl}

\address[ustsaddress]{School of Mathematical Science, Suzhou University of Science and Technology, Suzhou, PR China}

\address[xjtluaddress]{Department of Applied Mathematics, Xi'an Jiaotong-Liverpool University, Suzhou, PR China}

% \author[uvaaddress,itmoaddress]{Pavel Zun}
% \ead{p.zun@uva.nl}

% \address[uvaaddress]{Computational Science Lab, Institute for Informatics, Faculty of Science, University of Amsterdam, The Netherlands}

% \address[itmoaddress]{National Center for Cognitive Research, ITMO University, Saint Petersburg, Russia}

\begin{abstract}
Parameter estimation is a central task in data-driven learning of dynamical systems. It aims to recover the underlying physical parameters from observed time-series data, thereby providing interpretable insights into the physical mechanisms governing the system. Gradient/derivative matching methods based on Gaussian process provide an efficient way to perform parameter estimation. Those methods avoid repeated numerical integration and enforce local derivative consistency. However, such local matching may result in global inconsistency with the governing flow map, particularly under scarce and noisy observations. To address this limitation, we propose a framework based on Gaussian process learning with flow map refinement (GPL-FMR), a two-stage parameter estimation framework. The first stage is based on Gaussian process learning algorithm and the posterior obtained from which is transferred as an informative prior to the second stage based on flow-map refinement. The second stage further improves the parameter estimation via optimisation based on global dynamical constraints. We demonstrate and analyse its performance on multiple numerical examples, including the Van der Pol oscillator, the Lotka-Volterra model, and the Lorenz-63 system. The results show that the proposed framework consistently improves parameter estimation accuracy, particularly under scarce and noisy observations.

\end{abstract}


\end{frontmatter}

% \linenumbers
% \numberwithin{table}{section}
% \numberwithin{figure}{section}

\section{Introduction}
Data-driven learning of dynamical systems is one of the cutting-edge topics in scientific machine learning. The primary objective is to accurately infer the underlying system dynamics from observations, either in a continuous- or discrete-time formulation. It could subsequently serve as surrogate models in parametric scenarios, such as optimisation and control \cite{Bottcher2022,Zolman2025}, and uncertainty quantification \cite{Chu016,Tong2021,Cheng2023}. Alternatively, it can provide insights into the underlying mechanisms of the system by inferring physically meaningful parameters or identifying characteristic structural behaviours \cite{Prokop2026,GILPIN20201}. 

In general, data-driven learning can be categorised into three types depending on the purpose and prior knowledge of the system. The first type of learning assumes that no prior knowledge of the system is available and aims to uncover the underlying dynamical mechanism directly from data. Such scenarios are commonly referred as system identification \cite{Jin2012,Steven2016,Fattahi2018}. Sparse identification of nonlinear dynamics (SINDy) is one of the examples \cite{Steven2016}. The core idea of SINDy is to identify the active terms governing the dynamics by promoting sparsity from a candidate library of nonlinear functions of the states. Typically, the active terms and their associated coefficients identified through sparse regression retain physical interpretability and will facilitate subsequent analysis of the system dynamics. The second type of learning mainly focuses on approximating the dynamics within a prescribed space rather than identifying the actual representation of dynamics globally \cite{Chen2018neuralode,Greydanus2019,Rath2021,Schmid2022}. Such reparametrisation of the system depends on the chosen approximation paradigm. A representative example in continuous form is the NeuralODE, where the system dynamics are approximate by a neural network \cite{Chen2018neuralode}. The model parameters, i.e., weights and biases of nodes, are trained to adequately represent the dynamics through observed data. Those model parameters typically do not have any physical meaning unless constructed intentionally. Dynamics mode decomposition-based methods also fall in this category \cite{Schmid2022}. It bridges time-discrete solution sequences with linear transformation based on spectral bases. The last type of methods stem from scenarios where partial prior knowledge of the system is available, such as its governing structure, while the exact values of the physical parameters remain unknown. The problem consequently boils down to parameter estimation, aiming to infer the parameter values that best explain the observations \cite{beck1977parameter,Lillacci2010,GUO2022115336}. This setting is also closely related to inverse uncertainty quantification, where uncertain model parameters are inferred from observational data and their associated uncertainties are propagated to improve the predictive capability of the underlying model \cite{smith2024uncertainty}. 

Most data-driven approaches to parameter estimation, such as \cite{Chen2018neuralode,Schmid2022,GUO2022115336,Kramer2024}, calibrate the governing model through trajectory fitting. Specifically, the unknown parameters are determined by minimising the discrepancy between the observed states and the model-based trajectory obtained by numerical time integration, either within a deterministic optimisation framework or through a probabilistic formulation. Although trajectory fitting directly enforces agreement between the simulated and observed dynamics over time, it requires repeated numerical integration throughout the optimisation process. This can result in substantial computational cost, particularly for high-dimensional, stiff, or multiscale systems. On the other hand, gradient/derivative matching methods provide an alternative by replacing global trajectory fitting with a local comparison between the model vector field and the estimated time derivatives of the observed states \cite{Dondelinger2013,Steven2016,Wenk2019,YE2024108184}. Such methods can substantially reduce the computational burden associated with parameter estimation by avoiding repeated solution of the governing differential equations. Their performance, however, depends critically on the accuracy of the derivative estimates. Direct numerical differentiation, for example, finite difference approximation, amplifies observational noise and introduces discretisation errors. Furthermore, although derivative matching avoids explicit long-time integration, finite difference derivative estimates are still interpreted as local one-step discrete-time evolution approximations \cite{DAVID2023112495}. Consequently, developing a more stable and reliable alternative to direct numerical differentiation remains a central challenge in derivative matching-based parameter estimation.

Gaussian processes (GPs) provide a natural alternative to direct numerical differentiation. In regression, a GP places a prior over the unknown function specified by a mean function and a covariance kernel, and updates this prior by conditioning on the observed data \cite{williams2006gaussian}. The resulting posterior provides a nonparametric approximation of the underlying function and the quantification of the associated predictive uncertainty. A particularly useful property of GP is that a GP remains its Gaussianity under linear operations. Provided that the covariance kernel is sufficiently smooth, the derivative of the GP-smoothed trajectories can be obtained analytically through differentiation of the kernel. This property has motivated GP-based gradient/derivative matching methods, initially formulated through product-of-experts (PoE) constructions \cite{Calderhead2008,Wenk2019,Dondelinger2013}. In such formulations, one probabilistic expert represents the derivative distribution induced by the GP reconstruction, while the other represents the derivative implied by the governing differential equation for a given parameter set. Their product therefore favours parameter values that assign high probability under both derivative descriptions simultaneously. Ye et al. \cite{YE2024108184} proposed Gaussian process learning (GPL) within a Bayesian inference framework to address derivative estimation under sparse and noisy observations. Latent force models are also closely related \cite{Alvarez2009,Mauricio2013}, but they place GP priors on latent inputs, whereas GP derivative matching uses GPs to reconstruct the states and their derivatives. 

Despite these advantages, GP-based derivative matching still inherits the intrinsic limitation of derivative matching methods, i.e., it lacks the global dynamical constraint imposed by the ODE flow map. The method enforces only local consistency between the GP-estimated derivatives and the model vector field at a finite set of observations. Such pointwise agreement does not necessarily ensure that the reconstructed states lie on a single trajectory generated by the inferred differential equation. This issue is particularly pronounced when observations are scarce and noisy since the GP reconstruction and derivative estimates may then be strongly influenced by the bias-variance trade-off in such scenarios. Consequently, small local reconstruction errors can yield a low derivative matching residual but still produce a trajectory that is globally inconsistent with the observed dynamics.  

In this work, we propose a two-stage parameter estimation method, namely Gaussian process learning with flow map refinement (GPL-FMR). In the first stage, Gaussian process learning is employed to obtain a computationally efficient parameter estimate through probabilistic derivative matching. This estimate, together with its associated uncertainty, is subsequently transferred to the second stage as an informative prior. The second stage performs multi-shooting trajectory refinement through the ODE flow map, thereby enforcing the global dynamical constraint that is absent from the first stage. The informative initialisation provided by the first stage substantially improves the efficiency and robustness of the trajectory-based inference, while the flow map likelihood supports better parameter estimates. The proposed method is evaluated on several numerical examples, including the Van der Pol oscillator, the Lotka-Volterra model, and the Lorenz-63 system, to evaluate its performance against feature regression and GPL.

\section{Preliminary}
\subsection{Problem statement}
Consider an autonomous dynamical system,
\begin{equation}
\dot{\mathbf{x}}(t) = f(\mathbf{x}(t);\bm{\theta})\quad \mathrm{with} \quad \mathbf{x}(t_0) = \mathbf{x}_0    
\end{equation}
where $\mathbf{x}(t)= [x_1(t),x_2(t),\cdots,x_N(t)]\trp\in \mathbb{R}^N$ denotes the state vector of the system, $\mathbf{x}_0$ specifies the initial condition at $t=t_0$, and $f(\cdot;\bm{\theta})$ defines the vector field characterising the dynamics of the system parametrised by $\bm{\theta}\in\mathbb{R}^{N_p}$. A time series of states observations $\mathbf{X} = \{\mathbf{z}(t_i)\}_{i=1}^{N_t}$ is collected at time instances $\mathcal{T} =\{t_i\}_{i=1}^{N_t}$, with additive Gaussian white noise $\bm{\epsilon}\sim \mathcal{N}(0,\mathbf{R})$, i.e.,
\begin{equation}\label{eq:data_noise}
 \mathbf{z}(t_i) = \mathbf{x}(t_i) + \epsilon,    
\end{equation}
where $\mathbf{z}(t_i) = [z_1(t_i),\cdots,z_{N}(t_i)]^{\top}$, $\mathbf{R}=\operatorname{diag}(\bm\sigma_n^2)$ and $\bm{\sigma}_n^2 = [\sigma_{n,1}^2,\cdots,\sigma_{n,N}^2]^{\top} \in \mathbb{R}^{N}$ denotes the variance vector of the noise constructing the diagonal covariance matrix $\mathbf{R}$. Note that the observation can also be denoted component-wisely as $\{\mathbf{z}(t_i)\}_{i=1}^{N_t} = \{\mathbf{u}_i\}_{i=1}^N$, where $\mathbf{u}_i = [z_i(t_1),z_i(t_2),\cdots, z_i(t_{N_t})]^{\top}$. We aim to infer the unknown parameters $\bm\theta$ that lead to those observations with uncertainty quantification.
 
\subsection{Gaussian process and regression}
Here we present a general description of GP and its regression formulation. A GP can be viewed as a stochastic process that defines a distribution over functions. In particular, any finite collection of random variable $\{F(x_i)\}_{i=1}^{m}$ drawn from a GP $F(x)$ defined on a non-empty set $x\in\mathcal{X}$ with mean function $s(\cdot)$ and covariance function $k(\cdot,\cdot)$ for any finite set of $\{x_1,\cdots,x_m\} \in \mathcal{X}$ follows a joint Gaussian distribution,
\begin{equation}
\begin{bmatrix}
F(x_1)\\
\vdots\\
F(x_m)
\end{bmatrix}
\sim
\mathcal{N}\left(
\begin{bmatrix}
s(x_1)\\
\vdots\\
s(x_m)
\end{bmatrix},
\begin{bmatrix}
k(x_1,x_1) & \cdots & k(x_1,x_m)\\
\vdots & \ddots & \vdots\\
k(x_m,x_1) & \cdots & k(x_m,x_m)
\end{bmatrix}
\right).
\end{equation}
while the corresponding GP is denoted as,
\begin{equation}\label{eq:GP}
    F(\cdot) \sim \mathcal{GP}(s(\cdot),k(\cdot,\cdot)).
\end{equation}
The covariance function $k(\cdot,\cdot)$ is a positive definite kernel and the choice of which should reflect the properties of $F(x)$. For example, the squared-exponential kernel implies sample paths that are infinitely mean-square differentiable, whereas the smoothness of a Matérn kernel depends on its smoothness parameter \cite{williams2006gaussian}. 

In terms of regression tasks, consider $y\in\mathbb{R}$ is the response of a physical process or computer experiment $y = F(\bm{x}) + \epsilon$ with zero-mean Gaussian noise $\epsilon\sim \mathcal{N}(0,\sigma^2)$. The latent mapping $F(\bm{x})$ follows a GP prior as stated in Equation~\eqref{eq:GP} with zero mean and a defined covariance function, i.e.,
\begin{equation}\label{eq:PDE_GP}
    F(\cdot) \sim \mathcal{GP}\Bigl(0,k_{\bm{\phi}}(\cdot,\cdot)\Bigr),
\end{equation}
where $\bm{\phi}$ denotes the collection of hyperparameters from the chosen kernel function, including the noise variance $\sigma^2$. With observed data $(\mathbf{X},\mathbf{y})=\big\{(\bm{\bm{x}}_{i}, y_{i})\big\}_{i=1}^{N_u}$, those hyperparameters $\bm{\phi}$ that forms covariance matrix $\mathbf{K}$ can be estimated via minimising negative log marginal likelihood function,
\begin{equation}
\tilde{\bm{\phi}} = \arg\max_{\bm{\phi}} \log p(\mathbf{y}| \bm{\phi})= \frac{1}{2} \log |\bm{\mathrm{K}}| +\frac{1}{2}\mathbf{y}^{\top}\bm{\mathrm{K}}^{-1}\mathbf{y} +\frac{N_u}{2}\log 2\pi
\label{eq:pde_likelihood}
\end{equation}
where the covariance matrix element $[\bm{\mathrm{K}}]_{ij} = k_{\bm{\phi}}(\bm{x}_i,\bm{x}_j)+\sigma^2\delta(\bm{x}_i,\bm{x}_j)$. With the GP prior, a joint distribution between the existing data $\mathbf{y}$ and the prediction $F(\bm{x}^*)$ at a new point $\bm{x}^*$ can be constructed,
\begin{equation} 
\left[\begin{array}{c} \mathbf{y} \\ F(\bm{x}^*) \end{array}\right] \Big| ~\mathbf{X},\tilde{\bm{\phi}}
\sim 
\mathcal{N}\left(\mathbf{0}, \,\left[\begin{array}{cc}
 \bm{\mathrm{K}} & \mathbf{k}_*^{\top}\\
\mathbf{k}_* &  k_{\tilde{\bm{\phi}}}(\bm{x}^*,\bm{x}^*)
\end{array}\right]\right),
\end{equation}
where $\bm{\mathrm{k}}_*=k_{\tilde{\bm{\phi}}}\left(\bm{x}^*,\mathbf{X}\right)$. 
Therefore the corresponding predictive distribution of sample point $F(\bm{x}^*)$ can be achieved by conditioning on the observation $\mathbf{y}$,
\begin{equation} \label{eq:SGP_prediction}
     F(\bm{x}^*) | \, \mathbf{y},\mathbf{X},\tilde{\bm{\phi}} \sim \mathcal{N}\Bigl(\bar{F}(\bm{x}^*),v(\bm{x}^*)\Bigr)
\end{equation}
with mean function $\bar{F}(\bm{x}^*) = \bm{\mathrm{k}}_*\bm{\mathrm{K}}^{-1} \mathbf{y}$ and variance $ v_u(\bm{x}^*)=k_{\tilde{\bm{\phi}}}(\bm{x}^*, \bm{x}^*)-\bm{\mathrm{k}}_* \bm{\mathrm{K}}^{-1} \bm{\mathrm{k}}_*^{\top}$. 

\section{Two-stage parameter estimation with uncertainty quantification}
\begin{figure}
    \centering
    \includegraphics[width=\linewidth]{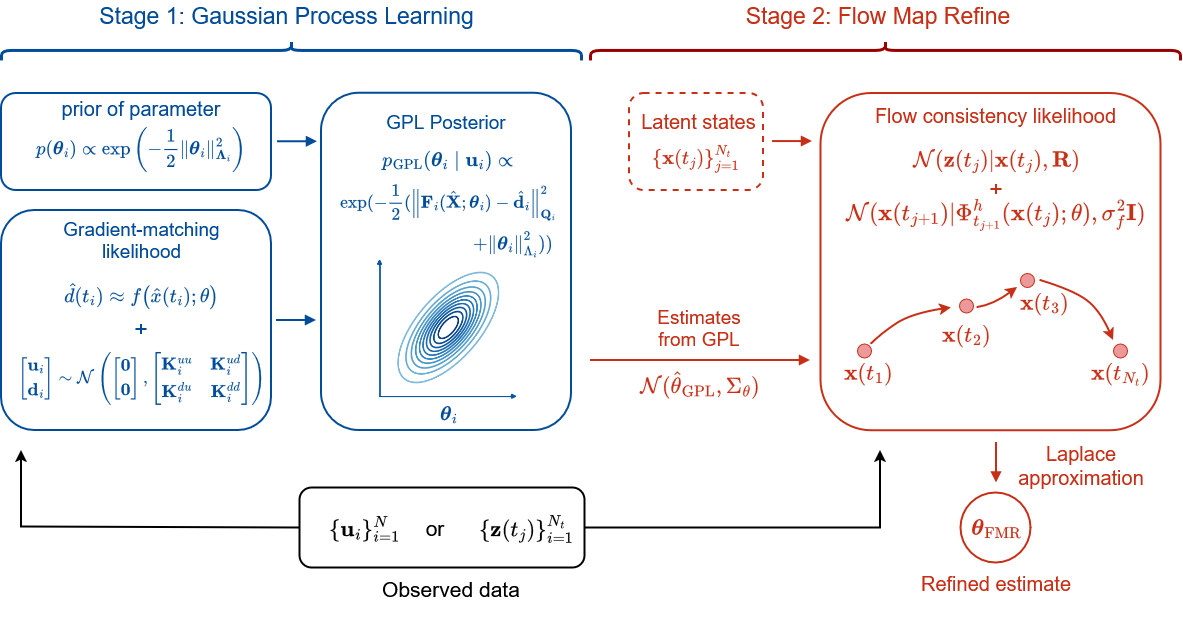}
    \caption{Schematic diagram of the proposed two-stage parameter estimation framework. The first stage employs Gaussian process learning (blue), while the second stage applies flow-map refinement (red) to further refine the parameter estimates.}
    \label{fig:diagram}
\end{figure}
A schematic diagram of the proposed two-stage parameter estimation method is demonstrated in Figure~\ref{fig:diagram}. In the first stage, Gaussian process learning is used to obtain an initial probabilistic estimate of the unknown parameters through a derivative-matching likelihood. This stage is computationally efficient and explicitly accounts for the uncertainty of the GP-induced derivative estimates. In the second stage, the GPL posterior is transferred as an informative empirical prior for a flow map refinement process. This stage constructs a likelihood with two complementary components, an observation model that links the latent states to the measured data, and a multi-shooting flow map model that enforces dynamical consistency between consecutive latent states. In this way, the proposed framework retains the computational efficiency of local derivative matching while refining its parameter estimates through observation fidelity and dynamical constraints imposed by the ODE flow map.

\subsection{Stage one: Gaussian process learning based on derivative matching likelihood}

Consider the $i$th state variable $x_i(t): \mathbb{R}\rightarrow\mathbb{R}$ with GP prior,
\begin{equation}
    x_i(t) \sim \mathcal{GP} (0, k_{\bm{\phi}_i}(\cdot,\cdot)), \quad i = 1,\cdots, N
\end{equation}
The core idea of Gaussian process learning stems from the fact that a GP is closed under linear operators. Therefore, the derivative process is jointly Gaussian with the original process provided the kernel is sufficiently differentiable,
\begin{equation}
\begin{bmatrix}
x_i(t) \\
\dot{x}_i(t)
\end{bmatrix} | \,\bm{\phi}_i
\sim
\operatorname{vec\text{-}\mathcal{GP}}
\left(
\begin{bmatrix}
0 \\
0
\end{bmatrix},
\begin{bmatrix}
k_{\bm{\phi}_i}(t,t') &
\partial_{t'}k_{\bm{\phi}_i}(t,t')
\\
\partial_tk_{\bm{\phi}_i}(t,t') &
\partial_t\partial_{t'}k_{\bm{\phi}_i}(t,t')
\end{bmatrix}
\right)
, \quad i = 1,\cdots, N.
\end{equation}
This vector-valued GP is parametrised by the hyperparameters \(\bm{\phi}_i\) of the kernel associated with the \(i\)th state variable. By further denoting time derivative $\mathbf{d}_i=\left[\dot{x}_i(t_1),\ldots,\dot{x}_i(t_{N_t})\right]^\top$ of trajectory $\mathbf{x}_i=\left[x_i(t_1),\ldots,x_i(t_{N_t})\right]^\top$ corresponding to the observations $\mathbf{u}_i=\left[u_i(t_1),\ldots,u_i(t_{N_t})\right]^\top$, the joint distribution of the discrete observation and the time derivative of its trajectory over time can be expressed as,
\begin{equation}
    \begin{bmatrix}
        \mathbf{u}_i\\
        \mathbf{d}_i
    \end{bmatrix}
    |\,
    \bm{\phi}_i
    \sim
    \mathcal{N}
    \left(
    \begin{bmatrix}
        \mathbf{0}\\
        \mathbf{0}
    \end{bmatrix},
    \begin{bmatrix}
        \mathbf{K}^{uu}_i & \mathbf{K}^{ud}_i\\
        \mathbf{K}^{du}_i & \mathbf{K}^{dd}_i
    \end{bmatrix}
    \right).
\end{equation}
where
\begin{equation*}
    \mathbf{K}^{uu}_i = k_{\bm{\phi}_i}(\mathcal{T},\mathcal{T}) + \sigma_{n,i}^2\mathbf{I},\quad\mathbf{K}^{ud}_i = \partial_{t'}k_{\bm{\phi}_i}(\mathcal{T},\mathcal{T}) = (\mathbf{K}^{du}_i)^{\top}, \quad \mathbf{K}^{dd}_i = \partial_t\partial_{t'}k_{\bm{\phi}_i}(\mathcal{T},\mathcal{T}).
\end{equation*}
Therefore the GP conditional mean of the time derivative and the corresponding variance that quantifies the uncertainty associated with the derivative estimate can be written as,
\begin{equation}
    p\left(
        \mathbf{d}_i
        \mid
        \mathbf{u}_i,\bm{\phi}_i,\sigma_{n,i}^2
    \right)
    =
    \mathcal{N}
    \left(
        \mathbf{K}^{du}_i(\mathbf{K}^{uu}_i)^{-1}\mathbf{u}_i,
        \mathbf{K}^{dd}_i -
         \mathbf{K}^{du}_i\left(\mathbf{K}^{uu}_i\right)^{-1}
    \mathbf{K}^{ud}_i
    \right),
    \label{eq:gp_derivative_conditional}
\end{equation}

To account for a possible discrepancy between the assumed vector field and the derivative of the latent trajectory, we introduce the probabilistic dynamical constraint,
\begin{equation}
    \mathbf{d}_i
    =
    \mathbf{F}_i(\hat{\mathbf{X}};\bm{\theta}_i)
    +
    \bm{\zeta}_i,
    \qquad
    \bm{\zeta}_i
    \sim
    \mathcal{N}
    \left(
        \mathbf{0},
        \gamma_i^2\mathbf{I}_{N_t}
    \right),
\label{eq:dynamical_discrepancy}
\end{equation}
where $\gamma_i^2$ denotes the dynamical-model discrepancy variance, and $\mathbf{F}_i(\hat{\mathbf{X}};\bm{\theta}_i)= [f_i\!\left(\hat{\mathbf{x}}(t_1);\bm{\theta}_i\right),\cdots, f_i\!\left(\hat{\mathbf{x}}(t_{N_t});\bm{\theta}_i\right)]^{\top}$ denotes the evaluation of the $i$th component of the vector field with regressed state $\hat{\mathbf{X}} =\{\hat{\mathbf{x}}(t_j)\}_{j=1}^{N_t}$ at the observation times. Consequently,
\begin{equation}\label{eq:ODE_likelihood}
    p\left(
        \mathbf{d}_i
        \mid
        \hat{\mathbf{X}},\bm{\theta}_i,\gamma_i
    \right)
    =
    \mathcal{N}
    \left(\mathbf{d}_i \mid
        \mathbf{F}_i(\hat{\mathbf{X}};\bm{\theta}_i),
        \gamma_i^2\mathbf{I}_{N_t}
    \right).
\end{equation}
Since $ \mathcal{N}\left(\mathbf{d}_i\mid\mathbf{F}_i(\hat{\mathbf{X}};\bm{\theta}_i),\gamma_i^2\mathbf{I}_{N_t}\right) = \mathcal{N}\left(\mathbf{F}_i(\hat{\mathbf{X}};\bm{\theta}_i)\mid\mathbf{d}_i,\gamma_i^2\mathbf{I}_{N_t}\right)$ as Gaussian distributions, integrating the product of \eqref{eq:gp_derivative_conditional} and \eqref{eq:ODE_likelihood} with respect to $\mathbf d_i$ gives,
\begin{align}
    p\left(
        \mathbf{F}_i(\hat{\mathbf{X}};\bm{\theta}_i)
        \mid
        \mathbf{u}_i,\bm{\phi}_i,\gamma^2_i,\sigma^2_{n,i}
    \right)
    \nonumber &=
    \int
    p\left(
        \mathbf{F}_i(\hat{\mathbf{X}};\bm{\theta}_i)
        \mid
        \mathbf{d}_i,\gamma^2_i
    \right)
    p\left(
        \mathbf{d}_i
        \mid
        \mathbf{u}_i,\bm{\phi}_i,\sigma^2_{n,i}
    \right)
    \,\mathrm{d}\mathbf{d}_i
    \nonumber\\
    &\quad =
    \mathcal{N}
    \left(
        \mathbf{F}_i(\hat{\mathbf{X}};\bm{\theta}_i)
        \mid
        \mathbf{K}^{du}_i(\mathbf{K}_i^{uu})^{-1}\mathbf{u}_i,
        \mathbf{K}^{dd}_i -
         \mathbf{K}^{du}_i\left(\mathbf{K}^{uu}_i\right)^{-1}
    \mathbf{K}^{ud}_i+\gamma_i^2\mathbf{I}_{N_t}.
    \right),
    \label{eq:gpl_likelihood}
\end{align}
Integrating out the time derivative $\mathbf{d}_i$ offers a marginal derivative-matching likelihood for $\bm{\theta}_i$, which measures the compatibility between the vector-field derivative $\mathbf{F}_i(\hat{\mathbf{X}};\bm{\theta}_i)$ and the GP-induced derivative distribution. 
Consider a prior distribution on dynamical system parameters $\bm{\theta}_i$,
\begin{equation}
     p(\bm{\theta}_i)
    \propto
    \exp
    \left(
        -\frac12
        \|\bm{\theta}_i\|_{\bm{\Lambda}_i}^2
    \right).
\end{equation}
Using the prior and the derivative-matching likelihood, the posterior distribution of $\bm{\theta}_i$ is,
\begin{align}
    p_{\mathrm{GPL}}(\bm{\theta}_i\mid \mathbf{u}_i,\bm{\phi}_i,\gamma^2_i,\sigma^2_{n,i})
    &\propto
    p(\bm{\theta}_i)
    p\left(
        \mathbf F_i(\hat{\mathbf X};\bm{\theta}_i)
        \mid
        \mathbf u_i,\bm{\phi}_i,\gamma^2_i,\sigma^2_{n,i}
    \right),\\
    % \nonumber
    % &\propto p(\bm{\theta}_i) \mathcal{N}
    % \left(
    %     \mathbf{F}_i(\hat{\mathbf{X}};\bm{\theta}_i)
    %     \mid
    %     \mathbf{K}^{du}_i(\mathbf{K}_i^{uu})^{-1}\mathbf{u}_i,
    %     \mathbf{K}^{dd}_i -
    %      \mathbf{K}^{du}_i\left(\mathbf{K}^{uu}_i\right)^{-1}
    % \mathbf{K}^{ud}_i+\gamma_i^2\mathbf{I}_{N_t}
    % \right) \\
    &\propto \exp \left( -\frac{1}{2}\left( \left\| \mathbf{F}_i(\hat{\mathbf{X}};\bm{\theta}_i) - \hat{\mathbf{d}}_i \right\|^2_{\mathbf{Q}_i} + \left\|\bm{\theta}_i \right\|^2_{\Lambda_i}\right) \right)
    \label{eq:fbgpl_theta_posterior}
\end{align}
where $\|\bullet\|_{\mathbf{Q}_i} =\sqrt{\bullet^{\top}\mathbf{Q}_i \bullet}$, $\mathbf{Q}_i = \left(\mathbf{K}^{dd}_i - \mathbf{K}^{du}_i\left(\mathbf{K}^{uu}_i\right)^{-1}\mathbf{K}^{ud}_i+\gamma_i^2\mathbf{I}_{N_t} \right)^{-1}$  and $\hat{\mathbf{d}}_i = \mathbf{K}^{du}_i(\mathbf{K}_i^{uu})^{-1}\mathbf{u}_i$. In the exponential of Equation~\eqref{eq:fbgpl_theta_posterior}, a regularised weighted least squares estimator of $\bm{\theta}_i$ with precision matrix $\mathbf{Q}_i$ is presented. The first term presents the data fidelity of how the vector field is compatible with time derivatives approximated by Gaussian process regression \cite{YE2024108184}, while the prior behaviour serves as a regularisation term. The precision matrix $\mathbf{Q}_i$ for the data fidelity term introduces the uncertainty of the derivative into the posterior estimation. 
\begin{remark}
Note that here we present the prior distributions on parameters $\bm{\theta}_i$ of each individual dimension of the state for notation simplicity. It is straightforward to adapt to the case where physical parameters are shared across different equations,
\begin{equation}
     p_{\mathrm{GPL}} \left(\bm{\theta}\mid \{\mathbf{u}_i\}_{i=1}^N\right) \propto  \exp \left( -\frac{1}{2}\sum_{i=1}^N \left( \left\| \mathbf{F}_i(\hat{\mathbf{X}};\bm{\theta}) - \hat{\mathbf{d}}_i \right\|^2_{\mathbf{Q}_i} + \left\|\bm{\theta} \right\|^2_{\Lambda_i}\right) \right)
\end{equation}
where $\bm{\theta}=\bm{\theta}_1\cup\cdots\cup\bm{\theta}_N$ denotes the collection of all physical parameters in the system. Here we drop parameters $\bm{\phi}_i,\gamma^2_i,\sigma^2_{n,i}$ for notation simplicity.
\end{remark}

For the linear-in-parameter scenarios, the $i$th component of the vector field, i.e., $\mathbf{F}_i(\mathbf{x};\bm{\theta}_i)= \mathbf{g}_i(\mathbf{x})^{\top}\bm{\theta}_i$, where $\mathbf{g}_i$ is a prescribed vector-valued function. Under the standard Gaussian prior $p(\bm{\theta}_i)\propto\exp\left(-\frac{1}{2}\|\bm{\theta}_i\|_2^2\right)$ and evaluating the vector field at the GP-reconstructed states $\hat{\mathbf{X}}$,
the posterior distribution of $\bm{\theta}_i$ can be written as,
\begin{equation}
    p_{\mathrm{GPL}}(\bm{\theta}_i\mid\mathbf{u}_i)\propto
    \exp\left(
        -\frac{1}{2}
        \left(
            \|\mathbf{G}_i\bm{\theta}_i-\hat{\mathbf{d}}_i\|_{\mathbf{Q}_i}^{2}
            +
            \|\bm{\theta}_i\|_2^2
        \right)
    \right)
\end{equation}
where $\mathbf{G}_i= [\mathbf{g}_i(\hat{\mathbf{x}}(t_1))\mid\cdots\mid
\mathbf{g}_i(\hat{\mathbf{x}}(t_{N_t}))]^{\top}$. This is essentially a quadratic w.r.t $\bm{\theta}_i$, hence the posterior follows a Gaussian distribution,
\begin{equation}
    p_{\mathrm{GPL}}(\bm{\theta}_i\mid\mathbf{u}_i)
    \sim
    \mathcal{N}(\bm{\mu}_{\mathrm{GPL},i},\bm{\Sigma}_{\mathrm{GPL},i}),
\end{equation}
where posterior covariance $\bm{\Sigma}_{\mathrm{GPL},i}=\left(\mathbf{G}_i^{\top}\mathbf{Q}_i\mathbf{G}_i+\mathbf{I}\right)^{-1},$ and posterior mean $\bm{\mu}_{\mathrm{GPL},i}=\bm{\Sigma}_{\mathrm{GPL},i}\mathbf{G}_i^{\top}\mathbf{Q}_i\hat{\mathbf{d}}_i$.
% The posterior mean therefore coincides with a regularised generalised least-squares estimator of $\bm{\theta}_i$.
For nonlinear model parametrisation, the resulting posterior distribution can be approximated using standard numerical techniques, such as Markov chain Monte Carlo \cite{Jones2022} or variational inference \cite{Blei03042017}.

\subsection{Stage two: Bayesian flow-map refinement}
The derivative-matching likelihood in the first stage provides a probabilistic estimate of the unknown physical parameters by comparing the vector field with the GP-induced derivative distribution at given time instances. However, derivative matching is a local-in-time construction and does not directly impose the time evolution of the dynamical system. Therefore, in the second stage, a Bayesian flow-map refinement procedure is conducted, aiming to further improve the estimation of physical parameters $\bm{\theta}=\bm{\theta}_1\cup\cdots\cup\bm{\theta}_N$ as well as latent states $\{\mathbf{x}(t_i)\}_{i=1}^N$ additionally. Note that the latent states are not the main quantity of interest here and can be considered as nuisance variables. The refinement is mainly based on the Bayesian inference with a likelihood induced by the ODE flow map with multi-shooting algorithms and priors inherited from GPL.

The ODE flow-map likelihood essentially performs trajectory fitting by comparing the observed states with the trajectory generated through numerical integration from a given initial condition over the observation window. Consequently, errors in the initial state, parameter estimates, or numerical integration will accumulate over time and result in a highly sensitive and poorly conditioned optimisation problem. To alleviate this issue, we adopt a multi-shooting strategy that partitions the trajectory into shorter integration intervals and introduces the latent states of each data as additional optimisation variables. Each local trajectory segment is therefore fitted over a shorter time horizon, reducing error accumulation and sensitivity to parameter perturbations, while continuity constraints are imposed between neighbouring segments to recover a globally consistent trajectory.

Let $t_j$ be an arbitrary observation time instance, and the corresponding unknown latent state is denoted by $\mathbf{x}(t_j)$.
For a given parameter vector $\bm{\theta}$ and initial state $\mathbf{x}(t_j)$, the multi-shooting of the dynamical system from time instance $t_j$ to $t_{j+1}$ can be written as, 
\begin{equation}
\mathbf{x}(t_{j+1})
=
\bm{\Phi}^h_{t_{j+1}}(\mathbf{x}(t_j);\bm{\theta}),
\qquad j=1,\ldots,N_t-1,
\end{equation}
where $\bm{\Phi}^h_{t_{j+1}}(\mathbf{x}(t_j);\bm{\theta})$ denotes the numerically approximation of the exact flow,
$$\bm{\Phi}_{t_{j+1}}(\mathbf{x}(t_j);\bm{\theta}) = \mathbf{x}(t_j) + \int_{t_j}^{t_{j+1}}f(\mathbf{x}(\tau);\bm{\theta}) \mathrm{d}\tau$$ 
from $t_j$ to $t_{j+1}$. The corresponding flow consistency model can be subsequently constructed by assuming $\mathbf{x}(t_{j+1})
=
\bm{\Phi}^h_{t_{j+1}}(\mathbf{x}(t_j);\bm{\theta})
+\bm{\xi}$, where $\bm{\xi}\sim \mathcal{N}(0,\sigma^2_f\mathbf{I})$. Here the $\bm{\xi}$ represents a soft relaxation of the multi-shooting continuity condition as well as numerical integration error. Together with Equation~\eqref{eq:data_noise}, the physical parameters and the latent states can be inferred simultaneously through the likelihood,
\begin{equation}
p\left( \{ \mathbf{z}(t_j)\}_{j=1}^{N_t}, \{\mathbf{x}(t_j)\}_{j=2}^{N_t}
\mid\bm{\theta},\{\mathbf{x}(t_j)\}_{j=1}^{N_t},\mathbf{R},\sigma^2_f
\right)
=
\prod_{j=1}^{N_t}
\mathcal{N}(\mathbf{z}(t_j)|\mathbf{x}(t_j),\mathbf{R})
\prod_{j=1}^{N_t-1}
\mathcal{N}
\left(
\mathbf{x}(t_{j+1})
\mid
\bm{\Phi}^h_{t_{j+1}}(\mathbf{x}(t_j);\bm{\theta}),
\bm{\sigma}^2_f\mathbf{I}
\right).
\label{eq:ODE_data_likelihood}
\end{equation}
The first term in the likelihood enforces consistency between the latent states and the observed data, while the second term enforces dynamical consistency between consecutive latent shooting states through the numerical flow map. It is important to note that $\mathbf{x}(t_j)$ appears as both observations and variables in the likelihood. This is because the latent states $\{\mathbf{x}(t_j)\}_{j=1}^{N_t}$ are all treated as unknown shooting variables, meanwhile the latent states at next time instances $\{\mathbf{x}(t_j)\}_{j=2}^{N_t}$ are also considered as the observation after flow mapping, which constraints the model with the continuity of the dynamics.

To inherit the information from GPL posterior,  Equation~\eqref{eq:fbgpl_theta_posterior} is used as an informative prior for the physical parameters in this stage. Similarly, a prior can be assigned to the latent states. Combining the priors and the flow-map likelihood, the posterior of the physical parameters and latent states is given by
\begin{equation}
p_{\mathrm{FMR}}\left(
\bm{\theta},\{\mathbf{x}(t_j)\}_{j=1}^{N_t}
\mid
\{\mathbf{z}(t_j)\}_{j=1}^{N_t}
\right)\propto
p_{\mathrm{GPL}}(\bm{\theta})
p(\mathbf{S})\prod_{j=1}^{N_t}
\mathcal{N}(\mathbf{z}(t_j)|\mathbf{x}(t_j),\mathbf{R})
\prod_{j=1}^{N_t-1}
\mathcal{N}
\left(
\mathbf{x}(t_{j+1})
\mid
\bm{\Phi}^h_{t_{j+1}}(\mathbf{x}(t_j);\bm{\theta}),
\bm{\sigma}^2_f\mathbf{I}
\right).
\label{eq:bayesian_flow_refinement_posterior}
\end{equation}
where $\mathbf{S}=[\mathbf{x}(t_1)^{\top},\cdots,\mathbf{x}(t_{N_t})^{\top}]^{\top}$ denotes the random vector that stack all latent shooting states.

\begin{remark}
When observations are noise-free, the latent states are fixed to the observations, and $\prod_{j=1}^{N_t}
\mathcal{N}(\mathbf{z}(t_j)|\mathbf{x}(t_j),\mathbf{R})$ in Equation~\eqref{eq:ODE_data_likelihood} will disappear. Parameter inference is then governed solely by the flow-map consistency term, $\prod_{j=1}^{N_t-1}
\mathcal{N}
\left(
\mathbf{x}(t_{j+1})
\mid
\bm{\Phi}^h_{t_{j+1}}(\mathbf{x}(t_j);\bm{\theta}),
\bm{\sigma}^2_f\mathbf{I}
\right)$, where the discrepancy is controlled by $\sigma^2_f$. In the limiting case $\sigma^2_f\rightarrow0$ and with an exact flow map, this likelihood reduces to Dirac measures supported on trajectories satisfying the governing dynamics exactly, and the posterior is consequently concentrated on parameter values that reproduce the trajectory passing the observations.
\end{remark}

The number of unknowns in \eqref{eq:bayesian_flow_refinement_posterior} is $N_p+N_t N$, where $N_p$ denotes the number of physical parameters, $N$ is the dimension of the state space and $N_t$ denotes the number of time instances where data are observed. In practice, when either the number of observations or the state dimension becomes large, approximating the full posterior distribution can be computationally prohibitive. We therefore seek the maximum a posteriori (MAP) estimate instead, reducing the Bayesian inference problem to an optimisation problem,
\begin{equation}
\left(
\bm{\theta}_{\mathrm{FMR}}^*,
\{\mathbf{x}^*(t_j)\}_{j=1}^{N_t}
\right)
=
\underset{
\bm{\theta},\,\{\mathbf{x}(t_j)\}_{j=1}^{N_t}
}{\operatorname{argmin}}\,
\mathcal{J}
\label{eq:map_optimisation}
\end{equation}
where  
\begin{equation}
\begin{aligned}
\mathcal{J}
&=
\frac{1}{2}
\sum_{j=1}^{N_t}
\left(
\mathbf{z}(t_j)-\mathbf{x}(t_j)
\right)^{\top}
\mathbf{R}^{-1}
\left(
\mathbf{z}(t_j)-\mathbf{x}(t_j)
\right)+
\frac{1}{2\sigma_f^2}
\sum_{j=1}^{N_t-1}
\left\|
\mathbf{x}(t_{j+1})
-
\bm{\Phi}^{h}_{t_{j+1}}
\left(
\mathbf{x}(t_j);\bm{\theta}
\right)
\right\|^2
\\
&\quad
+
\frac{1}{2}
\left(
\bm{\theta}-\bm{\mu}_{\mathrm{GPL}}
\right)^{\top}
\bm{\Sigma}_{\mathrm{GPL}}^{-1}
\left(
\bm{\theta}-\bm{\mu}_{\mathrm{GPL}}
\right)
+
\frac{1}{2}
\left(
\mathbf{S}-\hat{\mathbf{S}}_{\mathrm{GP}}
\right)^{\top}
\hat{\bm{\Sigma}}_{\mathbf{S}}^{-1}
\left(
\mathbf{S}-\hat{\mathbf{S}}_{\mathrm{GP}}
\right).
\end{aligned}
\label{eq:map_objective}
\end{equation}
Here, the vector $\hat{\mathbf{S}}_{\mathrm{GP}}$ denotes the prior mean vector of the latent shooting states, while $\hat{\bm{\Sigma}}_{\mathbf{S}}$ denotes the associated covariance. Note that Equation~\eqref{eq:bayesian_flow_refinement_posterior} includes the prior $p(\mathbf{S})$ to provide a general Bayesian formulation for the latent shooting states. In practice, one may take values from the GP-regressed state estimates evaluated in the GPL stage, or simply drop this term in optimisation to avoid propagating potential smoothing bias or miscalibrated uncertainty from GP regression into the FMR stage. Consequently, the final objective function is determined directly by the observation-fidelity and flow-map consistency terms, whereas the GPL posterior information is transferred only through the informative prior on the physical parameters. The precision matrices in Equation~\eqref{eq:map_objective} act as weighting matrices in the MAP optimisation. In particular, $\mathbf{R}^{-1}$ determines the contribution of the observation fidelity, whereas $\sigma_f^{-2}\mathbf{I}$ controls the contribution of the flow-map consistency. A larger precision assigns a stronger penalty to the corresponding residual, while a smaller precision permits greater deviation in accordance with the assumed uncertainty. 

Apart from MAP estimates, one can also quantify the local posterior uncertainty of the refined physical parameters with Laplace approximation \cite{geisser1990validity}. It only requires the local curvature of the negative log-posterior at the MAP estimate, and hence provides an efficient Gaussian approximation to the parameter posterior.

\begin{remark}
In the present formulation, the latent shooting states are regarded as nuisance variables since they are not primary quantities of interest. Their introduction serves mainly to decompose long-time numerical integration into a sequence of shorter local integrations, thereby reducing sensitivity to perturbations in the initial state and physical parameters. Consequently, their estimated values should not be interpreted as the final reconstructed trajectory, but rather as auxiliary variables that indicate how effectively local flow-map consistency is enforced across consecutive shooting intervals.
\end{remark}

% Consequently, the physical parameters may be estimated accurately even when the individual shooting states remain comparatively weakly identified or exhibit larger posterior uncertainty for scarce and noise scenarios. Treating the shooting states as nuisance variables is therefore reasonable when the principal objective of FMR is parameter refinement: the states provide the local flexibility required to accommodate observation noise and flow-map discrepancy, while their estimated values need not be interpreted as the final reconstructed trajectory. If accurate latent-state reconstruction is also required, additional state regularisation or simply more data is needed

\section{Numerical Experiments}
In this section, the performance of the proposed GPL-FMR method is evaluated on three numerical examples. We mainly focus on the results of parameter estimation, i.e., the relative errors of MAP estimates from the FMR stage and their corresponding standard deviation from Laplace approximation. To quantify the accuracy, the relative error is computed as,
\begin{equation}\label{eq:error}
    e = 100\% \times \left(\frac{\|\bm{\theta}_{\mathrm{GT}} -\bm{\theta}^*_{\mathrm{FMR}}\|_2^2}{\|\bm{\theta}_{\mathrm{GT}}\|_2^2}\right)^{1/2}\,
\end{equation}
where $\bm{\theta}_{\mathrm{GT}}$ denotes the parameter vectors of ground truth. We compare the performance of the proposed method with feature regression (FR) with trend filter and Gaussian process learning (GPL) itself. Apart from the physical parameter estimates, we also report the inferred auxiliary shooting states to present the complete MAP solution and compare them with the GP-smoothed trajectory in the GPL stage to illustrate how the flow map refinement improves dynamical consistency.

\subsection{Van der Pol oscillator}

\begin{table}[t]
\centering
\begin{adjustbox}{width=\textwidth}
\begin{tabular}{c|ccc|ccc|ccc}
\toprule
& \multicolumn{3}{c|}{$10\%$ noise level}
& \multicolumn{3}{c|}{$20\%$ noise level}
& \multicolumn{3}{c}{$30\%$ noise level}
\\
\midrule
$N_t$
& FR & GPL & GPL-FMR
& FR & GPL & GPL-FMR
& FR & GPL & GPL-FMR
\\
\midrule
20
&  1.16
& 4.65 $\pm~(2.87\times 10^{-2})$
& \textbf{3.12} $\pm~(7.18\times 10^{-3})$
& 1.19
& 5.44 $\pm~(2.72\times 10^{-2})$
& \textbf{3.25} $\pm~(7.70 \times 10^{-3})$
& 1.19
& 6.44 $\pm~(2.80\times 10^{-2})$
& \textbf{2.89} $\pm~(6.87\times 10^{-3})$
\\
\midrule
30
& 1.21
& 2.84 $\pm~(6.54\times10^{-2})$
& \textbf{3.05}$\pm~(6.97\times10^{-3})$
& 1.22
& 3.08 $\pm~(1.11\times 10^{-1})$
& \textbf{3.07} $\pm~(7.42\times 10^{-3})$
& 1.17
& \textbf{2.96} $\pm~(1.69 \times 10^{-1})$
& 3.17 $\pm~(7.59\times 10^{-3})$
\\
\midrule
40
& 1.02
& 2.30 $\pm~(5.60\times10^{-2})$
& \textbf{3.03} $\pm~(6.84\times 10^{-3})$
& 1.00
& 1.91 $\pm~(8.86\times 10^{-2})$
& \textbf{2.97} $\pm~(6.40 \times10^{-3})$
& 0.98
& 1.66 $\pm~(1.27 \times 10^{-1})$
& \textbf{2.98} $\pm~(6.36 \times 10^{-3})$
\\
\midrule
50
& 0.88
& 2.28 $\pm~(5.49\times 10^{-2})$
& \textbf{2.99} $\pm~(6.85\times 10^{-3})$
& 0.88
& 1.86 $\pm~(8.59 \times 10^{-2})$
& \textbf{2.99} $\pm~(6.73\times 10^{-3})$
& 0.88
& 1.65 $\pm~(1.19\times10^{-1})$
& \textbf{3.09} $\pm~(7.18 \times 10^{-3})$
\\
\midrule
100
& 1.28
& 3.16 $\pm~(4.71\times 10^{-2})$
& \textbf{3.01} $\pm~(5.29\times 10^{-3})$
& 1.15
& 2.26 $\pm~(7.10 \times 10^{-2})$
& \textbf{3.02} $\pm~(5.26\times 10^{-3})$
& 1.02
& 1.78 $\pm~(7.92 \times 10^{-2})$
& \textbf{3.01} $\pm~(5.14 \times 10^{-3})$
\\
\bottomrule
\end{tabular}
\end{adjustbox}
\caption{Mean $\pm$ standard deviation of the estimated Van der Pol oscillator parameter under $10\%$, $20\%$, and $30\%$ noise levels with $N_t=20$ and $N_t=100$ observations}. 
\label{tab:vdp_parameter}
\end{table}

\begin{figure}[t!]
    \centering
    \includegraphics[width=\linewidth]{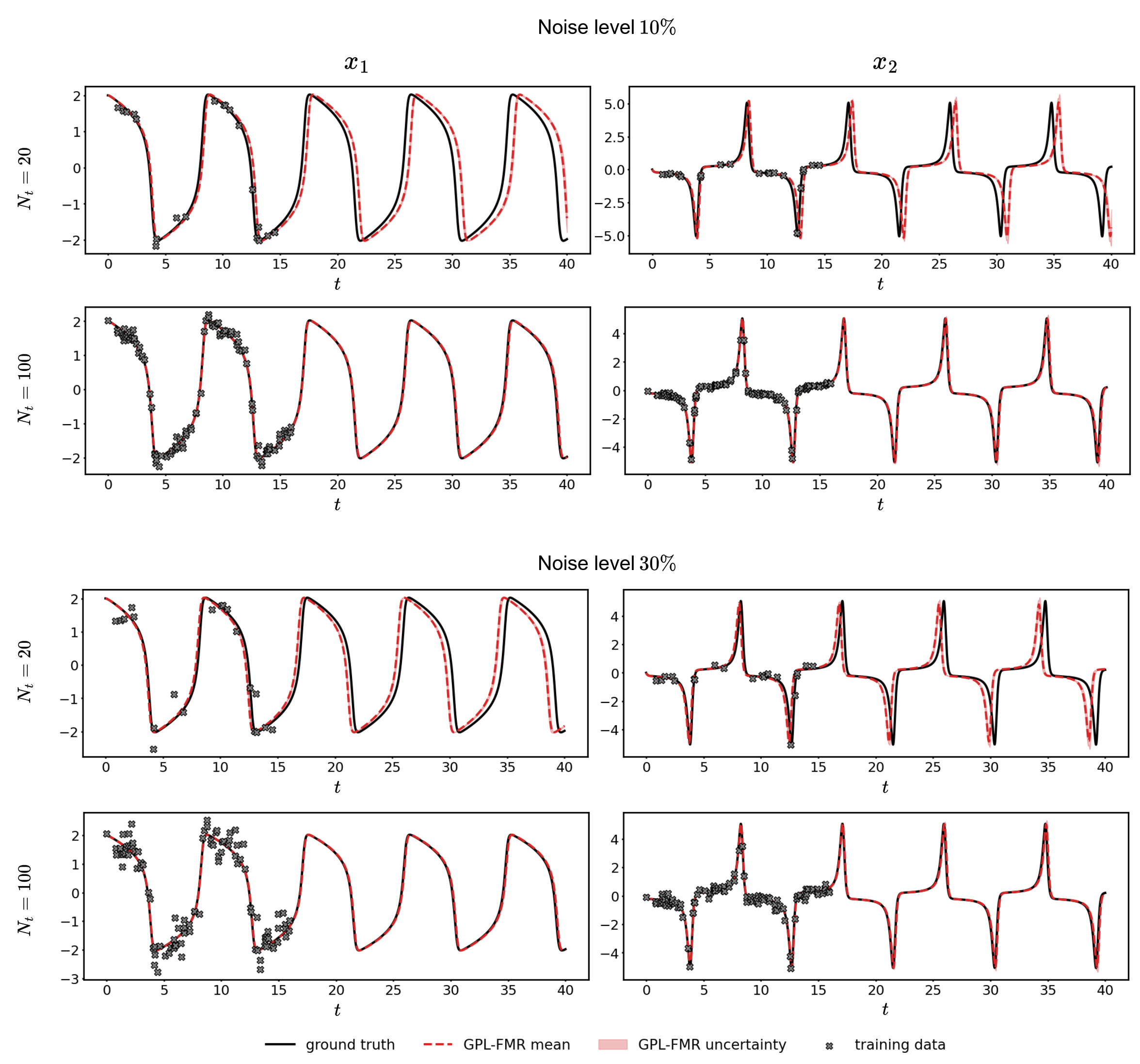}
    \caption{Trajectory reconstruction for the Van der Pol oscillator using the estimated parameters and the prescribed initial condition, with $N_t=20$ and $N_t=100$ observations under $10\%$ and $30\%$ noise levels. The uncertainty bands correspond to two standard deviations obtained from the Laplace approximation.}
    \label{fig:vdp_uq}
\end{figure}

\begin{figure}
    \centering
    \includegraphics[width=\linewidth]{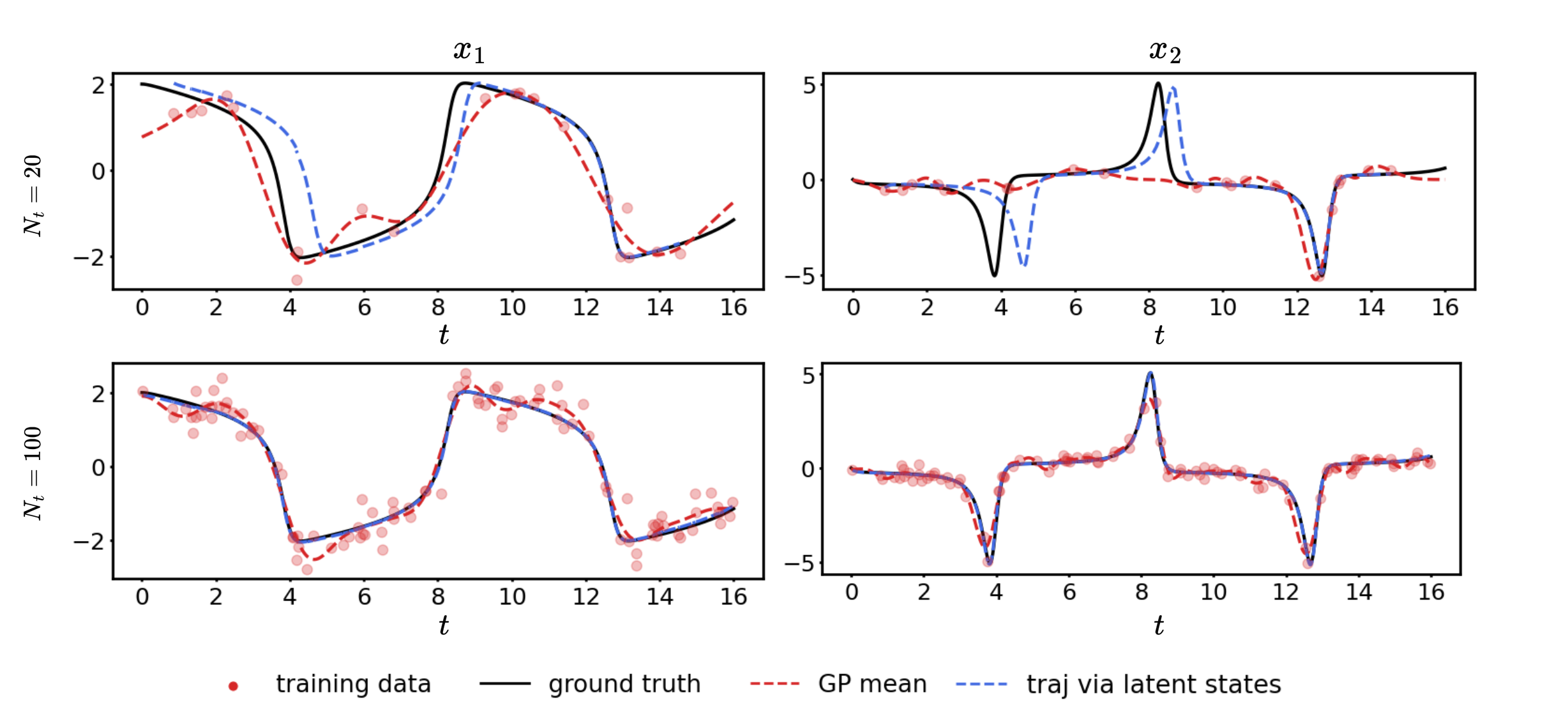}
    \caption{Comparison between the trajectories reconstructed from the inferred latent states and the GP-smoothed trajectories obtained from the GPL stage for the Van der Pol oscillator under 30\% noise.}
    \label{fig:vdp_latent}
\end{figure}

Van der Pol oscillator describes a non-conservative, oscillating system with nonlinear damping \cite{Pol01111926}. It is a second-order differential equation and can be written as,
\begin{equation}
\begin{cases}
\dot{x}_1 = x_2\\
\dot{x}_2 = \mu(1-x_1^2)x_2-x_1    
\end{cases}
\end{equation}
where $x_1$ and $x_2$ denote the position and velocity of the system, respectively. The parameter $\mu$ controls the strength of nonlinear damping. For a small $\mu$, the stable limit cycle has an amplitude of approximately 2, independent of the initial condition after the transient has disappeared. When $\mu$ increases, the waveform becomes more asymmetric and non-sinusoidal with a relatively slow build-up process and a rapid release. We test our proposed algorithm on the $\mu=3$ scenario. 

The trajectory of the system is generated from numerical integration via an implicit multi-step variable-order method \cite{Shampine1997} with a discrete time-step size $\mathrm{d}t=10^{-3}$ over $[0,T]$ where $T=40$. The initial condition is chosen as $x_1(0) = 2$ and $x_2(0) =0 $. Subsequently, the observational data are randomly chosen from the trajectory. In order to test the performance of GPL-FMR, we test multiple scenarios with the number of observational data increasing from 20 to 100 gradually. The experiments are conducted under $10\%$, $20\%$ and $30\%$ noise levels. The noise is assumed to follow a zero-mean Gaussian distribution with its standard deviation specified as the prescribed noise level multiplied by the mean magnitude of the data. The variances in Equation~\eqref{eq:map_objective} are set to be $\sigma_f^2 = 1\times 10^{-6}$ and $\mathbf{R} = \sigma_n^2\mathbf{I}$ where $\sigma_n^2=1\times 10^{-4}$. Note that the variances here behave as weights for each term in the optimisation rather than actual calibration of the data and flow variances.

The comparison of parameter estimation results using three methods is demonstrated in Table~\ref{tab:vdp_parameter}. The best result in each case is highlighted in bold. Overall, the proposed GPL-FMR outperforms the other two methods in nearly all test cases, with only one exception occurring at $N_t = 30$ under 30\% noise level. The GPL-FMR remains capable of recovering $\mu$ close to the ground truth value even under challenging conditions, such as 30\% noise with only 20 observations. In contrast, feature regression performs poorly in most cases. Its estimated values deviate substantially from the ground truth $\mu=3$ and cluster around 1. GPL provides reasonable estimates when the noise level is moderate and sufficient observations are available. However, its performance deteriorates as the noise level increases and becomes less stable when the observations are scarce.

Apart from mean estimates, the standard deviation obtained from the Laplace approximation is also shown in Table~\ref{tab:vdp_parameter}. A substantial reduction in posterior uncertainty can be observed after the flow-map refinement. While the GPL estimates exhibit standard deviations of approximately $10^{-1}$ to $10^{-2}$, the corresponding GPL-FMR uncertainties are consistently reduced to the order of $10^{-3}$ across different data sizes and noise levels. Moreover, the GPL uncertainty generally increases as the observation noise becomes stronger, whereas the GPL-FMR uncertainty remains comparatively stable. 
% This indicates that the flow-map likelihood introduces additional curvature in the local posterior geometry and more strongly constrains the parameter around the MAP estimate, leading to a considerably more concentrated local posterior. However, it is important to point out that these standard deviations characterise the local uncertainty implied by the Laplace approximation and therefore do not necessarily capture non-Gaussianity or multimodality of the full posterior distribution.

The trajectories reconstructed by the inferred parameters and the prescribed initial condition $x_1(0) = 2,\,x_2(0) =0 $ are shown in Figure~\ref{fig:vdp_uq}. The red dashed line denotes the reconstructed trajectories, while the black solid lines are the ground truth. The noise-corrupted data are presented in grey cross markers. As one can see, the reconstructed trajectories agree closely with the ground truth. For the scenarios with $N_t=20$, slight deviation gradually emerge over time under both 10\% and 30\% noise. In contrast, the inferred parameter reproduces the ground truth trajectory with high accuracy when $N_t = 100$.

The latent states jointly inferred with the physical parameters in Equation~\eqref{eq:map_objective} are also visualised in Figure~\ref{fig:vdp_latent}. The blue curves denote trajectories generated by evolving the system from the inferred latent states using the estimated physical parameters, while the red dashed curves show the GP-smoothed trajectories obtained from the GPL stage, and the red markers denote the observations under 30\% noise. For the sparse case with $N_t=20$, neither reconstruction fully agrees with the ground truth. Nevertheless, the trajectories based on the inferred latent states preserve the overall dynamical pattern except that a slight temporal shift remains. In contrast, the GP-smoothed trajectories fail to capture several local features of the true dynamics. For example, the peak behaviour of $x_2$ around $t=4$ is flattened. When $N_t=100$, both reconstructions improve substantially. The trajectories generated from the inferred latent states closely follow the ground truth, whereas the GP-smoothed trajectories still miss some detailed dynamical features.

\subsection{Lotka-Volterra model}

\begin{table}[t!]
\centering
\begin{adjustbox}{width=\textwidth}
\begin{tabular}{c|ccc|ccc|c}
\toprule

% =========================================================
% 10% noise level
% =========================================================
\multicolumn{8}{c}{$10\%$ noise level} \\
\midrule

& \multicolumn{3}{c|}{$N_t = 20$}
& \multicolumn{3}{c|}{$N_t = 100$}
& \\
\midrule

Parameters
& FR 
& GPL
& GPL-FMR
& FR
& GPL
& GPL-FMR
& GT \\
\midrule

$\alpha$
& 1.03
& 1.35 $\pm~(1.05 \times 10^{-1})$
& \textbf{1.54} $\pm~(3.77 \times 10^{-3})$
& \textbf{1.50}
& 1.46 $\pm~(3.53 \times 10^{-2})$
& 1.48 $\pm~(1.70 \times 10^{-3})$
& 1.5 \\
\midrule

$\beta$
& 1.03
& \textbf{1.01} $\pm~(8.40 \times 10^{-2})$
& \textbf{0.99} $\pm~(2.54 \times 10^{-3})$
& 1.02
& 0.97 $\pm~(2.20 \times 10^{-2})$
& \textbf{1.00} $\pm~(1.07 \times 10^{-3})$
& 1 \\
\midrule

$\sigma$
& 1.16
& 1.28 $\pm~(5.80 \times 10^{-2})$
& \textbf{0.99} $\pm~(2.71 \times 10^{-3})$
& 0.90
& 0.93 $\pm~(1.50 \times 10^{-2})$
& \textbf{1.01} $\pm~(1.19 \times 10^{-3})$
& 1 \\
\midrule

$\rho$
& 2.17
& 3.21 $\pm~(1.29 \times 10^{-1})$
& \textbf{2.94} $\pm~(7.59 \times 10^{-3})$
& 2.65
& 2.79 $\pm~(4.64 \times 10^{-2})$
& \textbf{3.04} $\pm~(3.90 \times 10^{-3})$
& $3$ \\

% =========================================================
% 20% noise level
% =========================================================
\midrule
\multicolumn{8}{c}{$20\%$ noise level} \\
\midrule

& \multicolumn{3}{c|}{$N_t = 20$}
& \multicolumn{3}{c|}{$N_t = 100$}
& \\
\midrule

Parameters
& FR
& GPL
& GPL-FMR
& FR
& GPL
& GPL-FMR
& GT \\
\midrule

$\alpha$
& 1.17
& 1.25 $\pm~(1.88 \times 10^{-1})$
& \textbf{1.60} $\pm~(3.98 \times 10^{-3})$
& \textbf{1.53}
& 1.39 $\pm~(7.01 \times 10^{-2})$
& \textbf{1.47} $\pm~(1.73 \times 10^{-3})$
& 1.5 \\
\midrule

$\beta$
& 1.20
& 0.93 $\pm~(1.52 \times 10^{-1})$
& \textbf{0.97} $\pm~(2.52 \times 10^{-3})$
& 1.04
& 0.92 $\pm~(4.32 \times 10^{-2})$
& \textbf{1.00}$ \pm~(1.11 \times 10^{-3})$
& 1 \\
\midrule

$\sigma$
& \textbf{1.00}
& 1.24 $\pm~(1.08 \times 10^{-1})$
& 0.99 $\pm~(2.65 \times 10^{-3})$
& 0.82
& 0.86 $\pm~(2.81 \times 10^{-2})$
& \textbf{1.02} $\pm~(1.23 \times 10^{-3})$
& 1 \\
\midrule

$\rho$
& 1.80
& \textbf{2.90} $\pm~(2.41 \times 10^{-1})$
& 2.87 $\pm~(7.28 \times 10^{-3})$
& 2.45
& 2.58 $\pm~(8.70 \times 10^{-2})$
& \textbf{3.07} $\pm~(4.01 \times 10^{-3})$
& $3$ \\

% =========================================================
% 30% noise level
% =========================================================
\midrule
\multicolumn{8}{c}{$30\%$ noise level} \\
\midrule

& \multicolumn{3}{c|}{$N_t = 20$}
& \multicolumn{3}{c|}{$N_t = 100$}
& \\
\midrule

Parameters
& FR
& GPL
& GPL-FMR
& FR
& GPL
& GPL-FMR
& GT \\
\midrule

$\alpha$
& 1.24
& 1.15 $\pm~(3.05 \times 10^{-1})$
& \textbf{1.68} $\pm~(4.30 \times 10^{-3})$
& \textbf{1.52}
& 1.31 $\pm~(1.07 \times 10^{-1})$
& 1.47 $\pm~(1.80 \times 10^{-3})$
& 1.5 \\
\midrule

$\beta$
& 1.27
& 0.89 $\pm~(2.49 \times 10^{-1})$
& \textbf{0.97} $\pm~(2.55 \times 10^{-3})$
& 1.03
& 0.87 $\pm~(6.54 \times 10^{-2})$
& \textbf{1.02} $\pm~(1.19 \times 10^{-3})$
& 1 \\
\midrule

$\sigma$
& 0.80
& \textbf{1.02} $\pm~(1.67 \times 10^{-1})$
& \textbf{0.98} $\pm~(2.59 \times 10^{-3})$
& 0.74 
& 0.79 $\pm~(4.00 \times 10^{-2})$
& \textbf{1.03} $\pm~(1.27 \times 10^{-3})$
& 1 \\
\midrule

$\rho$
& 1.41
& 2.12 $\pm~(3.45 \times 10^{-1})$
& \textbf{2.80} $\pm~(7.00 \times 10^{-3})$
& 2.24
& 2.40 $\pm~(1.25 \times 10^{-1})$
& \textbf{3.08} $\pm~(4.13 \times 10^{-3})$
& $3$ \\
\bottomrule
\end{tabular}
\end{adjustbox}
\caption{Mean $\pm$ standard deviation of the estimated Lotka-Volterra model's parameters under $10\%$, $20\%$, and $30\%$ noise levels with $N_t=20$ and $N_t=100$ observations.}
\label{tab:lv_parameter}
\end{table}

\begin{figure}[t!]
    \centering
    \includegraphics[width=\linewidth]{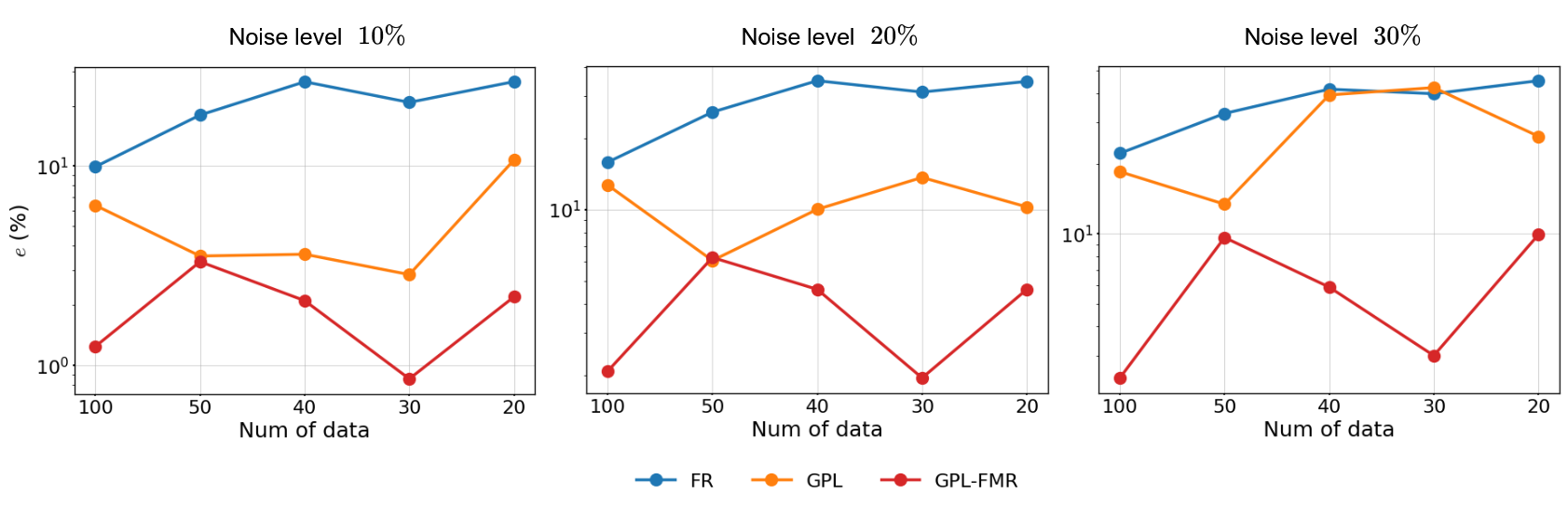}
    \caption{Comparison of the relative errors in parameter estimation for the Lotka-Volterra model using feature regression (FR), GPL, and the proposed GPL-FMR method, with 20,30,40,50, and 100 observations under 10\%, 20\%, and 30\% noise levels.}
    \label{fig:lv_error}
\end{figure}

\begin{figure}[t!]
    \centering
    \includegraphics[width=\linewidth]{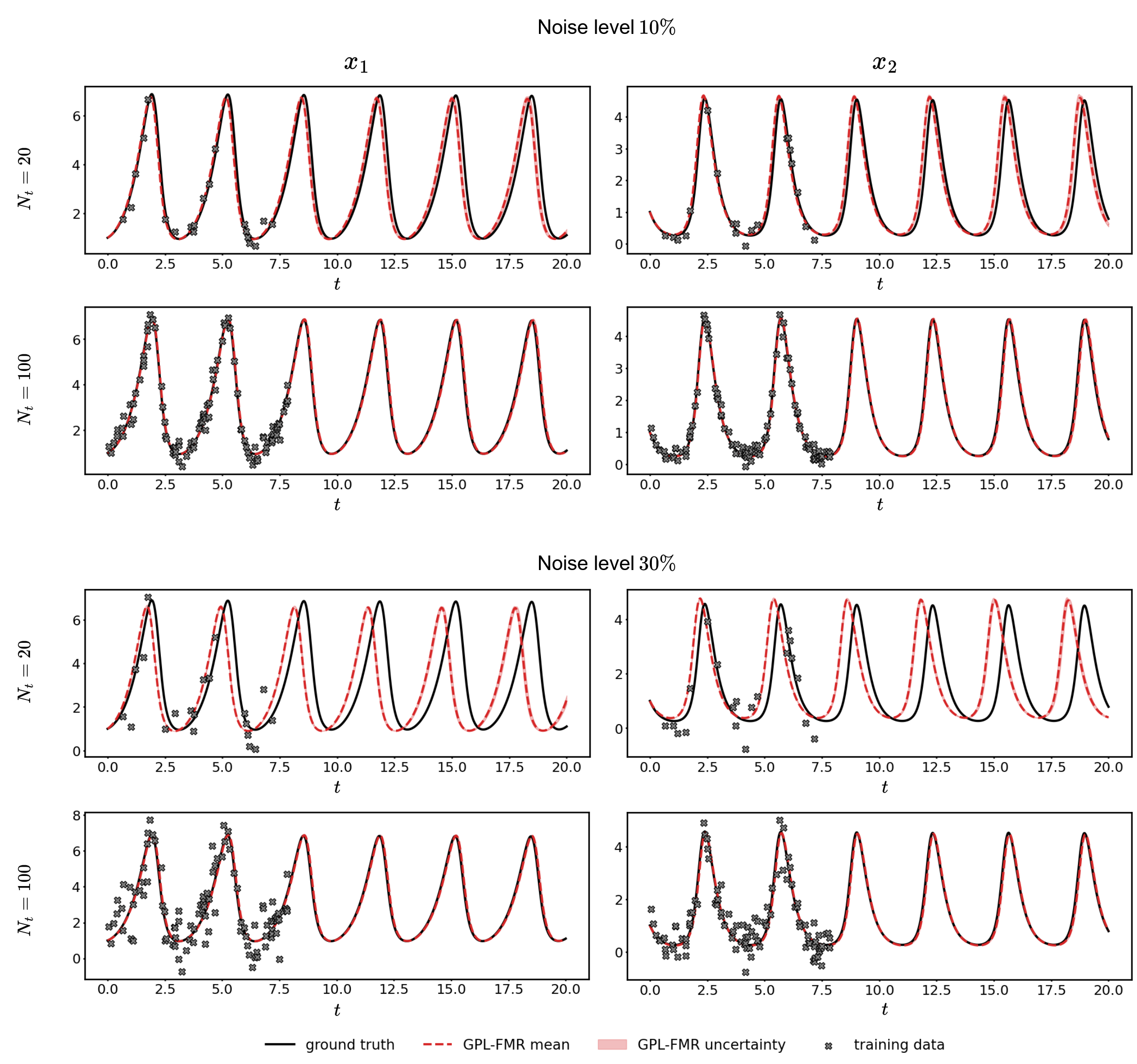}
    \caption{Trajectory reconstruction for the Lotka-Volterra model using the estimated parameters and prescribed initial condition, with $N_t=20$ and $N_t=100$ observations under $10\%$ and $30\%$ noise levels. The uncertainty bands correspond to two standard deviations obtained from the Laplace approximation.}
    \label{fig:lv_traj}
\end{figure}

\begin{figure}[th!]
    \centering
    \includegraphics[width=\linewidth]{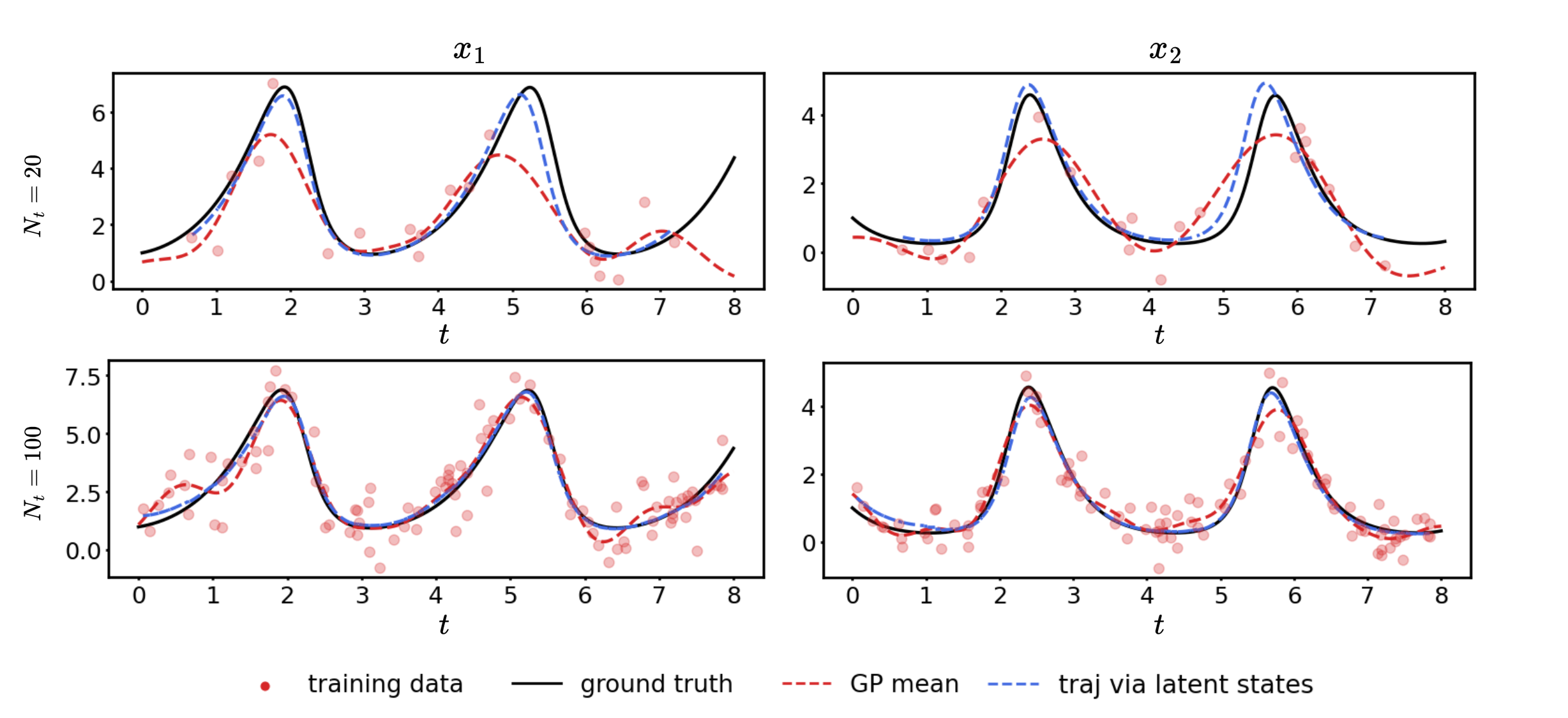}
    \caption{Comparison between the trajectories reconstructed from the inferred latent states and the GP-smoothed trajectories obtained from the GPL stage for the Lotka-Volterra model under 30\% noise.}
    \label{fig:lv_latent}
\end{figure}

The two-dimensional Lotka-Volterra equation models the population dynamics between two species in an ecological system. The system can be written as,
\begin{equation}
\begin{cases}
\dot{x}_1=\alpha x_1-\beta x_1 x_2 \,,\\
\dot{x}_2=\delta x_1 x_2-\rho x_2 \,, 
\end{cases}
\end{equation}
where $x_1$ and $x_2$ denote the population states of two species respectively. The parameters $\alpha$ and $\rho$ govern the growth rate of the species, while $\beta$ and $\delta$ characterise the interaction between the two species. In this example, the system parameters are set to be $[\alpha, \beta, \delta, \rho] = [1.5,1,1,3]$. Similar to the Van der Pol oscillator, the observation of the system is generated from numerical integration from a prescribed initial condition $x_1(0) = 1$ and $x_2(0)=1 $ via an implicit multi-step variable-order method. The discrete time step size is $\mathrm{d}t=10^{-3}$ and the entire time window is $[0,T]$ where $T=20$. The same scenarios (20 to 100 observations as well as 10\% 20\% and 30\% noise levels) are set to 
test the performance of the proposed method.

Table~\ref{tab:lv_parameter} compares the parameter estimates obtained by three methods for the Lotka-Volterra model under 20 and 100 observations and 10\% and 30\% noise levels. Among the three methods, GPL-FMR again provides the most consistently accurate estimates across the four physical parameters. The improvement is particularly evident for $\rho$, where both FR and GPL exhibit substantial bias as the noise level increases. GPL-FMR on the other hand remains much closer to the ground truth. A similar improvement is observed for $\beta$ and $\sigma$. For $\alpha$, GPL-FMR also performs well in most of the cases, with a few exceptions that are slightly outperformed by another method. Moreover, Figure~\ref{fig:lv_error} visualises the relative errors of all estimated parameters, computed using Equation~\eqref{eq:error}. The blue, orange, and red curves represent the results obtained by feature regression, GPL, and GPL-FMR, respectively. Consistent with the results reported in Table~\ref{tab:lv_parameter}, GPL-FMR consistently achieves the lowest relative errors across most test scenarios. In the test case with $N_t = 50$, the GPL performance is comparable to GPL-FMR. Meanwhile, GPL also shows a significant advantage over the feature regression method.

The GPL posterior standard deviations range approximately from $10^{-2}$ to $10^{-1}$ and generally increase with the observation noise, whereas the GPL-FMR standard deviations obtained from the Laplace approximation remain consistently of order $10^{-3}$. Increasing the number of observations from $N_t=20$ to $N_t=100$ further reduces the GPL-FMR uncertainty for all parameters. 

The reconstructed trajectories using the estimated parameters and the prescribed initial condition in Figure~\ref{fig:lv_traj} for the Lotka-Volterra model exhibit close agreement with the ground truth across the different noise levels. For $N_t=20$, small deviations gradually accumulate over time, whereas for $N_t=100$, the trajectories reconstructed using the inferred parameters remain highly consistent with the ground-truth dynamics. Besides, the trajectories generated from the inferred latent states using the estimated physical parameters reproduce the underlying dynamics accurately for both $N_t=20$ and $N_t=100$, as shown in Figure~\ref{fig:lv_latent}. In contrast, the GP-smoothed trajectories from the GPL stage fail to recover the true oscillatory behaviour reliably, particularly in the case with less data. This indicates that the flow-map refinement provides latent states that are more dynamically consistent with the governing system than the GP regression alone.

\subsection{Lorenz-63 Model}

\begin{table}[t!]
\centering
\begin{adjustbox}{width=\textwidth}
\begin{tabular}{c|ccc|ccc|c}
\toprule

% =========================================================
% 10% noise level
% =========================================================
\multicolumn{8}{c}{$10\%$ noise level} \\
\midrule

& \multicolumn{3}{c|}{$N_t = 20$}
& \multicolumn{3}{c|}{$N_t = 100$}
& \\
\midrule

Parameters
& FR
& GPL
& GPL-FMR
& FR
& GPL
& GPL-FMR
& GT \\
\midrule

$\sigma$
& 1.04
& 3.63 $\pm~(1.84\times 10^{0})$
& \textbf{9.38} $\pm~(2.91\times 10^{-3})$
& 5.97
& 8.93 $\pm~(1.33\times 10^{-1})$
& \textbf{10.06} $\pm~(2.43\times 10^{-3})$
& 10 \\
\midrule

$\rho$
& 31.20
& \textbf{28.16} $\pm~(9.47\times 10^{-1})$
& 28.28 $\pm~(2.12\times10^{-3})$
& 26.31
& \textbf{27.97} $\pm~(4.75\times 10^{-2})$
& 27.90 $\pm~(1.03\times 10^{-3})$
& 28 \\
\midrule

$\beta$
& 1.45
& 2.31 $\pm~(7.99\times 10^{-2})$
& \textbf{2.73} $\pm~(4.12\times 10^{-4})$
& 1.80
& \textbf{2.66} $\pm~(1.42\times10^{-2})$
& 2.65 $\pm~(2.01\times 10^{-4})$
& 8/3 \\

% =========================================================
% 20% noise level
% =========================================================
\midrule
\multicolumn{8}{c}{$20\%$ noise level} \\
\midrule

& \multicolumn{3}{c|}{$N_t = 20$}
& \multicolumn{3}{c|}{$N_t = 100$}
& \\
\midrule

Parameters
& FR
& GPL
& GPL-FMR
& FR
& GPL
& GPL-FMR
& GT \\
\midrule

$\sigma$
& 1.05
& 3.41 $\pm~(1.90\times 10^{0})$
& \textbf{9.66} $\pm~(3.18\times 10^{-3})$
& 5.38
& 7.51 $\pm~(1.50\times 10^{-1})$
& \textbf{10.18} $\pm~(2.39\times 10^{-3})$
& 10 \\
\midrule

$\rho$
& 29.93
& 27.35 $\pm~(1.13\times 10^{0})$
& \textbf{28.13} $\pm~(2.09\times 10^{-3})$
& 25.96
& 26.71 $\pm~(6.58\times 10^{-2})$
& \textbf{27.78} $\pm~(1.03\times 10^{-3})$
& 28 \\
\midrule

$\beta$
& 1.43
& 2.43 $\pm~(7.39\times 10^{-2})$
& \textbf{2.79} $\pm~(4.23\times 10^{-4})$
& 1.92
& 2.01 $\pm~(9.48\times 10^{-2})$
& \textbf{2.63} $\pm~(2.00\times 10^{-4})$
& 8/3 \\

% =========================================================
% 30% noise level
% =========================================================
\midrule
\multicolumn{8}{c}{$30\%$ noise level} \\
\midrule

& \multicolumn{3}{c|}{$N_t = 20$}
& \multicolumn{3}{c|}{$N_t =100$}
& \\
\midrule

Parameters
& FR
& GPL
& GPL-FMR
& FR
& GPL
& GPL-FMR
& GT \\
\midrule

$\sigma$
& 0.91
& 2.77$ \pm~(1.85 \times 10 ^0)$
& \textbf{9.73} $\pm~(3.14\times 10^{-3})$
&  4.37
&6.10 $\pm~(9.97 \times 10^{-1})$
& \textbf{10.45} $\pm~(2.45\times10^{-3})$
& 10 \\
\midrule

$\rho$
& 30.34
& 28.78 $\pm~(2.20 \times 10^{0})$
& \textbf{28.01} $\pm~(2.08\times 10^{-3})$
& 25.76
& 26.86 $\pm~(3.24 \times 10^{-1})$
& \textbf{27.66} $\pm~(1.03\times 10^{-3})$
& 28 \\
\midrule

$\beta$
& 1.43
& 1.89 $\pm~(1.01\times 10^0)$
& \textbf{2.90} $\pm~(4.36 \times 10^{-4})$
& 2.01
& 1.54 $\pm~(1.27\times 10^{-1})$
& \textbf{2.60} $\pm~(1.97\times 10^{-4})$
& 8/3 \\

\bottomrule
\end{tabular}
\end{adjustbox}

\caption{Mean $\pm$ standard deviation of the estimated Lorenz-63 model's parameters under $10\%$, $20\%$, and $30\%$ noise levels with $N_t=20$ and $N_t=100$ observations.}
\label{tab:Lorenz_parameters}
\end{table}

\begin{figure}[t!]
    \centering
    \includegraphics[width=\linewidth]{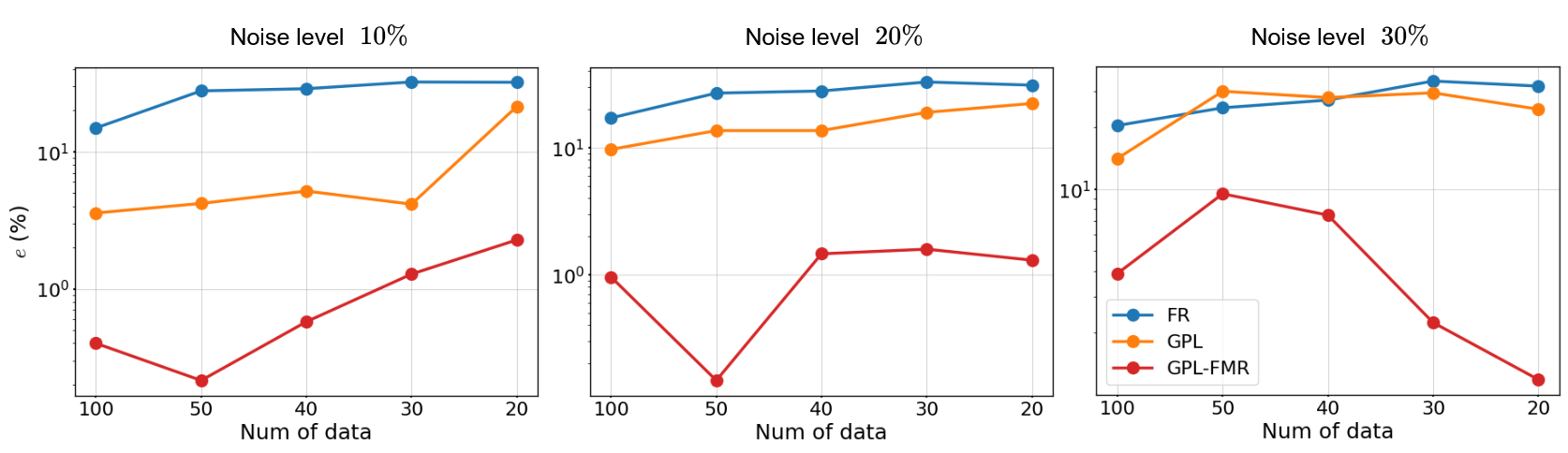}
    \caption{Comparison of the relative errors in parameter estimation for the Lotka-Volterra model using feature regression (FR), GPL, and the proposed GPL-FMR method, with 20,30,40,50, and 100 observations under 10\%, 20\%, and 30\% noise levels.}
    \label{fig:l63_error}
\end{figure}

\begin{figure}[t!]
    \centering
    \includegraphics[width=\linewidth]{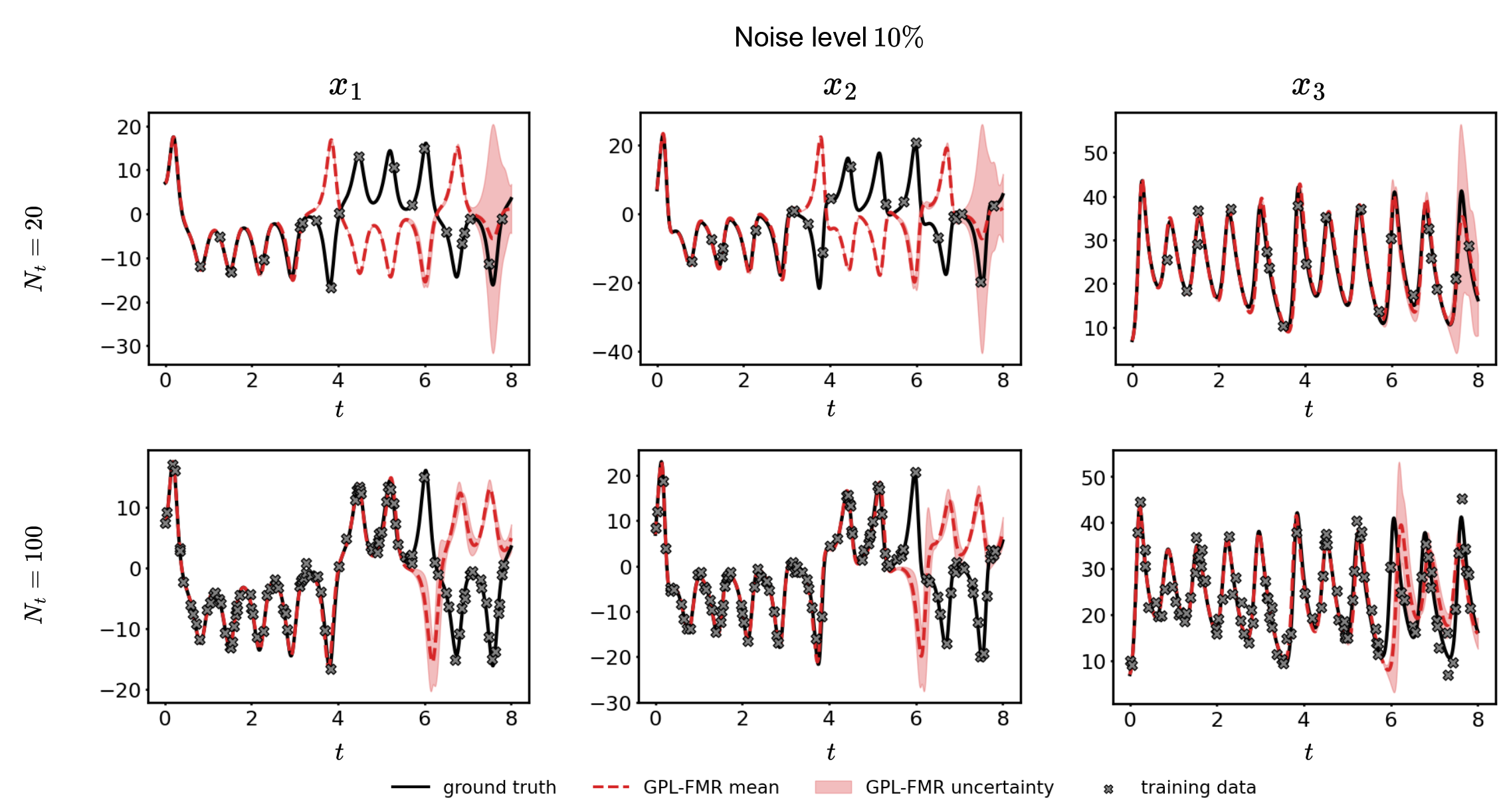}
    \caption{Trajectory reconstruction for the Lorenz-63 model using the estimated parameters and prescribed initial condition, with $N_t=20$ and $N_t=100$ observations under $10\%$ noise levels. The uncertainty bands correspond to two standard deviations obtained from the Laplace approximation.}
    \label{fig:l63_uq1}
\end{figure}

\begin{figure}[t!]
    \centering
    \includegraphics[width=\linewidth]{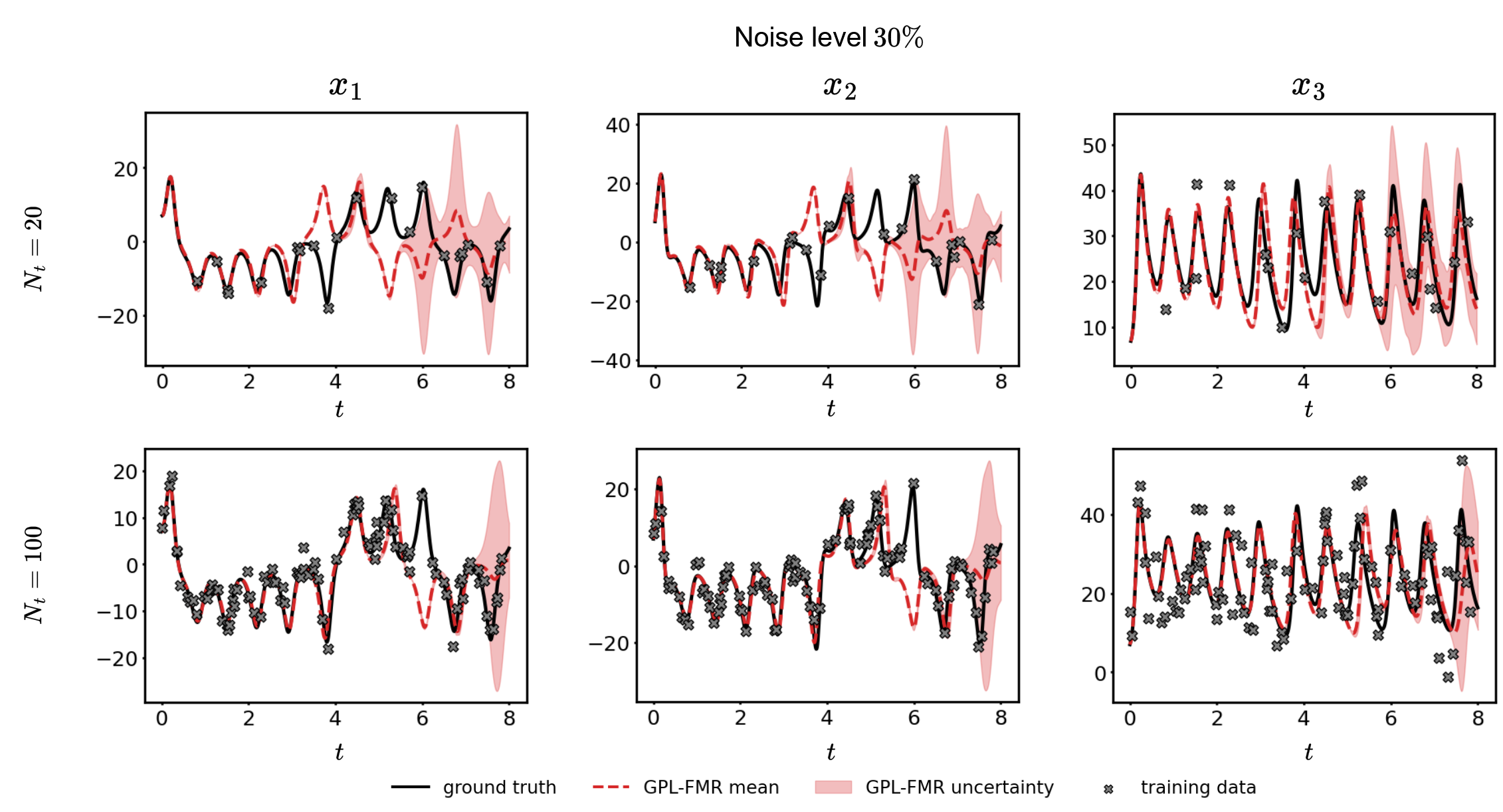}
    \caption{Trajectory reconstruction for the Lorenz-63 model using the estimated parameters and prescribed initial condition, with $N_t=20$ and $N_t=100$ observations under $30\%$ noise levels. The uncertainty bands correspond to two standard deviations obtained from the Laplace approximation.}
    \label{fig:l63_uq2}
\end{figure}

\begin{figure}[t!]
    \centering
    \includegraphics[width=\linewidth]{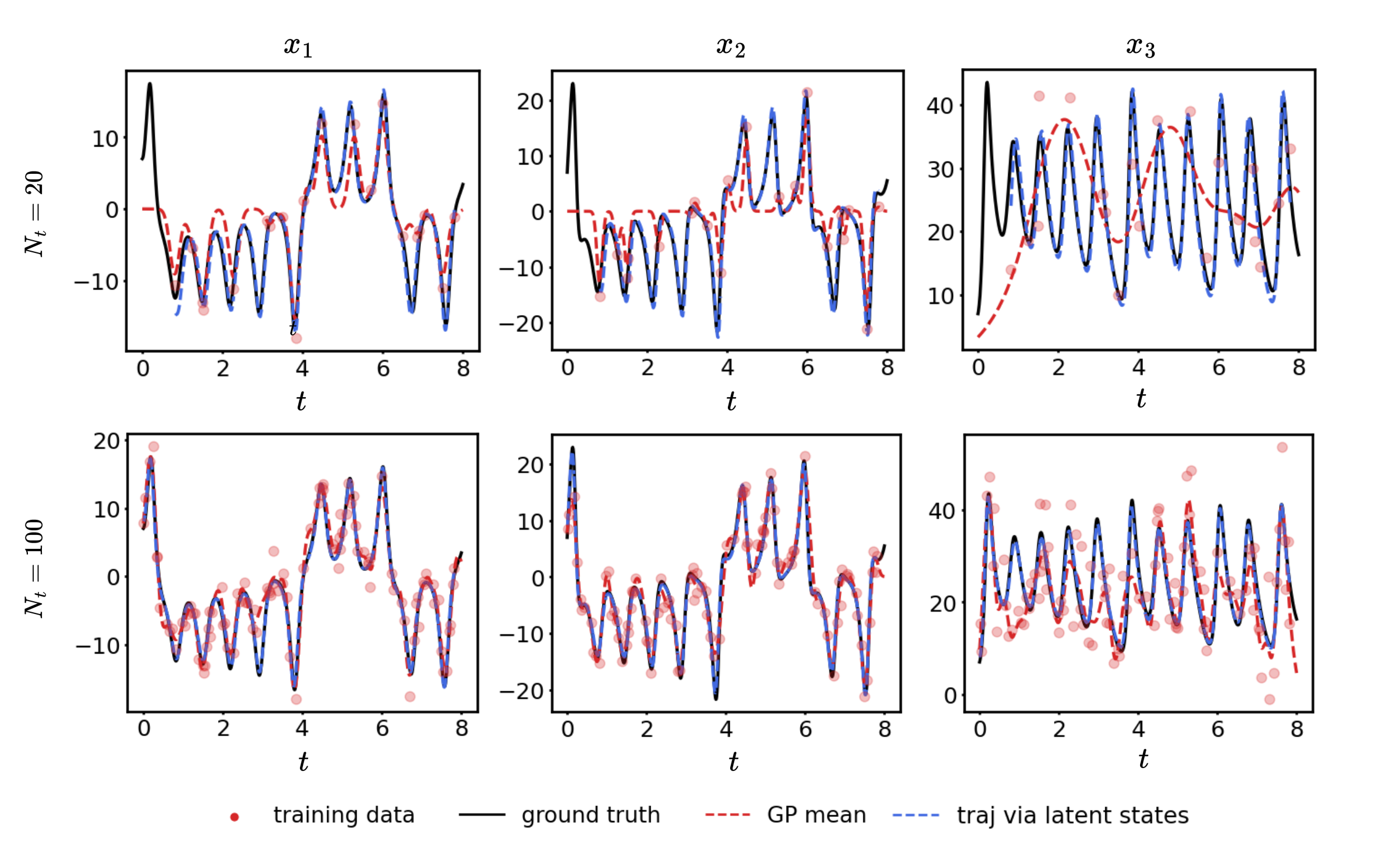}
    \caption{Comparison between the trajectories reconstructed from the inferred latent states and the GP-smoothed trajectories obtained from the GPL stage for the Lorenz-63 system under 30\% noise.}
    \label{fig:l63_latent}
\end{figure}

The Lorenz-63 model is a classic numerical example of dynamical systems, originally introduced as a simplified model of atmospheric convection. The behaviour of the state variables $x_1$, $x_2$, and $x_3$ are defined as,
\begin{align}
\begin{cases}
\dot{x}_1 = \sigma(x_2-x_1),\\
\dot{x}_2 = x_1(\rho-x_3)-x_2,\\
\dot{x}_3 = x_1x_2-\beta x_3,
\end{cases}
\end{align}
where $\sigma$, $\rho$, and $\beta$ are the physical parameters governing the dynamics, representing the Prandtl number, the Rayleigh number, and a geometric factor related to the fluid layer, respectively. We consider the standard parameter setting $\sigma=10$, $\rho=28$, and $\beta=8/3$, under which the system exhibits chaotic behaviour. In this regime, the trajectory evolves around two unstable lobes and is highly sensitive to perturbations in both the initial condition and the model parameters. The Lorenz-63 system therefore provides a challenging benchmark for assessing whether GPL-FMR can recover dynamically consistent parameter estimates from scarce and noisy observations.

The reference trajectory of the Lorenz-63 system is generated with a time-step size of $dt=10^{-2}$ over the interval $[0,T]$, where $T=8$. The initial condition is prescribed as $x_1(0)=7$, $x_2(0)=7$, and $x_3(0)=7$. Gaussian noise is subsequently added to construct observation scenarios with $10\%$, $20\%$, and $30\%$ noise levels. 

The parameter estimation results with $N_t=20$ and $N_t=100$ observations under $10\%$ and $30\%$ noise are reported in Table~\ref{tab:Lorenz_parameters}. Both feature regression and GPL are noticeably affected by observational noise. Their estimation accuracy deteriorates as the noise level increases. GPL generally outperforms feature regression and provides reasonably accurate estimates at $10\%$ noise level, but its performance degrades substantially under $20\%$ and $30\%$ noise scenarios. In contrast, the proposed GPL-FMR consistently yields more accurate parameter estimates across the tested cases, even with the challenging setting of only $20$ observations and $30\%$ noise.

We also present a more detailed comparison of the estimation errors in Figure~\ref{fig:l63_error}, where results are reported for $N_t=20$, $30$, $40$, $50$, and $100$. Overall, the proposed GPL-FMR method substantially outperforms both feature regression and GPL, with the parameter discrepancies reduced by approximately one order of magnitude in most cases. It can be observed that the GPL-FMR error does not decrease monotonically as the number of observations increases, particularly under the $30\%$ noise level. This behaviour can be attributed to the strongly anisotropic parameter sensitivity of the Lorenz-63 system. Different parameter perturbations affect the trajectory to different extents, and therefore a better trajectory reconstruction does not necessarily correspond to a uniformly smaller error in every parameter. As shown in Figure~\ref{fig:l63_uq2}, the reconstructed trajectory for $N_t=100$ under $30\%$ noise agrees more closely with the ground truth than that obtained with $N_t=20$, despite the absence of a corresponding monotonic reduction in the overall parameter error as shown in Figure~\ref{fig:l63_error}. Since the MAP objective in Equation~\eqref{eq:map_objective} is driven by trajectory and flow-map discrepancies rather than the parameter error itself, the optimisation naturally favours parameter combinations that best reproduce the observed dynamics. Nevertheless, GPL-FMR consistently provides more accurate parameter estimates than feature regression and GPL across the considered scenarios.

The trajectory reconstructions obtained from the estimated parameters with a prescribed initial condition are presented in Figures~\ref{fig:l63_uq1} and~\ref{fig:l63_uq2} for $10\%$ and $30\%$ noise levels respectively. For the $N_t=20$ cases, the reconstructed trajectories agree well with the ground truth up to approximately $t=3$, after which noticeable deviations emerge. This behaviour is expected for the Lorenz-63 system because its chaotic dynamics are highly sensitive to perturbations in both the physical parameters and the initial state. The agreement improves substantially when $N_t=100$. The reconstructed trajectories remain close to the ground truth up to approximately $t=5.5$. Among the three state variables, the reconstruction of $x_3$ is generally more accurate due to its comparatively smoother temporal behaviour.

Moreover, Figure~\ref{fig:l63_latent} shows that the trajectories generated from the jointly inferred latent states and physical parameters accurately reproduce the underlying chaotic dynamics for both $N_t=20$ and $N_t=100$ under 30\% noise. In contrast, the GP-smoothed trajectories fail to capture local dynamical features, including the oscillatory transitions and rapid variations characteristic of the Lorenz attractor. A particularly evident example can be observed for $x_3$ when $N_t=20$, where the GP over-smooths the data completely. Remarkably, even in the sparse case with $N_t=20$, the jointly inferred latent states recover these features with high fidelity. This demonstrates that the proposed flow-map refinement remains effective for chaotic systems.

\section{Discussion}
The proposed GPL-FMR framework combines GP-based derivative matching with the dynamical consistency imposed by the ODE flow map. The numerical results indicate that the flow map refinement plays an essential role in correcting the limitations inherited from the GPL stage. Under scarce and noisy observations, the GP-smoothed states and their derivative estimates can deviate substantially from the underlying dynamics, as illustrated in Figure~\ref{fig:vdp_latent}, Figure~\ref{fig:lv_latent} and Figure~\ref{fig:l63_latent}. Those poor estimates therefore influence the performance of parameter estimation of derivative matching methods. The FMR stage in the proposed GPL-FMR jointly refines the physical parameters and latent shooting states to constrain the inference through both observation fidelity and local flow-map consistency. The reconstructed trajectories further demonstrate that the inferred latent states provide a more dynamically consistent representation than the GP-smoothed states, especially when the latter fail to resolve important local features of the dynamics. The use of multi-shooting is also critical in this setting. Instead of propagating a single initial condition over the entire observation window, the trajectory is decomposed into shorter integration intervals, which reduces sensitivity to perturbations in the initial state and physical parameters. Moreover, conditional on the shooting states and parameters, the numerical integrations over individual intervals are independent and can therefore be evaluated in parallel, which provides an additional computational advantage during optimisation.

Although the formulation of FMR is Bayesian, approximating the full joint posterior of the physical parameters and latent shooting states through sampling methods such as MCMC can be computationally expensive. The dimension of the inference problem grows with the number of observations and state variables, while each posterior evaluation additionally requires numerical integration over all shooting intervals. The Laplace approximation therefore provides a practical compromise. After obtaining the MAP estimate, the local curvature of the negative log-posterior is used to construct a Gaussian approximation around the posterior mode. The resulting covariance characterises the local uncertainty of the inferred parameters and reflects how strongly the posterior constrains different parameter directions. This interpretation differs from sampling-based Bayesian inference, which characterises the full posterior distribution and can represent skewness, multimodality, and other non-Gaussian structures. Therefore, the approximation may become inaccurate when the posterior is strongly non-Gaussian, and may consequently underestimate uncertainty in highly nonlinear problems. Nevertheless, in the present framework, the Laplace approximation provides an effective and computationally efficient approach to uncertainty quantification.

\section{Conclusion}
This work proposed a two-stage framework for parameter estimation in dynamical systems from scarce and noisy observations. The first stage employs Gaussian process learning to obtain an efficient probabilistic parameter estimate through local derivative matching, while the second stage introduces a multi-shooting Bayesian flow-map refinement that jointly infers the physical parameters and latent shooting states. The GPL posteriors are incorporated as informative empirical priors, allowing the second stage to refine the estimates. To reduce the computational cost associated with full posterior estimation, MAP is employed, and the corresponding parameter uncertainty is quantified through a Laplace approximation. Numerical experiments on dynamical systems show that GPL-FMR generally improves the accuracy and robustness of parameter estimation over feature regression and the original GPL, particularly under sparse and noisy observations. These results demonstrate the benefit of combining the computational efficiency of derivative matching with the global dynamical information provided by the ODE flow map.

\section*{Acknowledgement}
D.Ye acknowledge the supports of Research Development Funding from Xi’an Jiaotong-Liverpool University under agreement RDF-25-01-009.

\section*{Data availability}
All the data and source codes to reproduce the results in this study are available on GitHub at \url{https://github.com/yhao9s9/gpl-fmr}.

\bibliographystyle{vancouver}
\bibliography{Reference}
% \section*{Appendix}
% Consider a reference domain $\Omega$ and a target domain $\Omega^*$ with distinct configurations. Let $n$ and $n^*$ be the the outward unit normals of each configuration, respectively. We assume that there existing a time-dependent diffeomorphism $\mathcal{X}(\bm{X},t)$ that give a point-to-point differentiable mapping from the reference domain to the target domain. The corresponding deformation gradient/Jacobian matrix $\bm{J}$ can be written as,
% \begin{equation}
%  \bm{J} = \frac{\partial\bm{x}}{\partial\bm{X}}
% \end{equation}
% and $\left|\bm{J}\right| = \mathrm{det}(\bm{J})$ denotes the determinant of the Jacobian matrix. The infinitesimal vector $d\bm{L}$ in $\Omega$ and $d\bm{l}$ in $\Omega^*$ is connected by $d\bm{l} = \bm{J}d\bm{L}$. Similarly the element volume between two domains are related by $dv = \left|\bm{J}\right| dV$, where  
\end{document}